\documentclass[twoside,11pt]{article}

\usepackage[preprint]{jmlr2e}
\hypersetup{%
    colorlinks=false,
    allbordercolors={1 1 1},
    hidelinks
}

\usepackage[dvipsnames]{xcolor}
\definecolor{kit-blue100}{cmyk}{.8,.5.,0,0}
\definecolor{kit-blue70}{cmyk}{.56,.35,0,0}
\definecolor{kit-blue50}{cmyk}{.4,.25,0,0}
\definecolor{kit-blue30}{cmyk}{.24,.15,0,0}
\definecolor{kit-blue15}{cmyk}{.12,.075,0,0}

\definecolor{kit-green100}{cmyk}{1,0,.6,0}
\definecolor{kit-green70}{cmyk}{.7,0,.42,0}
\definecolor{kit-green50}{cmyk}{.5,0,.3,0}
\definecolor{kit-green30}{cmyk}{.3,0,.18,0}
\definecolor{kit-green15}{cmyk}{.15,0,.09,0}

\newcommand{\code}{\texttt}
\usepackage{xspace}  
\newcommand{\sofatemplates}{\texttt{sofa\textunderscore{}templates}\xspace}
\newcommand{\sofazoo}{\texttt{sofa\textunderscore{}zoo}\xspace}
\newcommand{\sofaenv}{\texttt{sofa\textunderscore{}env}\xspace}
\newcommand{\sofagodot}{\texttt{sofa\textunderscore{}godot}\xspace}
\newcommand*{\eg}{\emph{e.g.}\@\xspace}

\newcommand*{\ie}{\emph{i.e.}\@\xspace}
\newcommand*{\etc}{\emph{etc.}\@\xspace}

\usepackage[acronym]{glossaries}
\newacronym{rl}{RL}{reinforcement learning}
\newacronym{rals}{RALS}{robot-assisted laparoscopic surgery}
\newacronym{ppo}{PPO}{Proximal Policy Optimization}
\newacronym{dvrk}{dVRK}{daVinci Research Kit}
\newacronym{sofa}{SOFA}{Simulation Open Framework Architecture}
\newacronym{pbd}{PBD}{position based dynamics}
\newacronym{fem}{FEM}{finite element method}
\newacronym{fls}{FLS}{Fundamentals of Laparoscopic Surgery TM}
\newacronym{gui}{GUI}{graphical user interface}
\newacronym{rcm}{RCM}{remote center of motion}

\usepackage{bm}      
\usepackage{booktabs} 
\usepackage{tabularx} 
\usepackage{ltablex} 
\newcolumntype{b}{X} 
\newcolumntype{s}{>{\hsize=.5\hsize}X} 

\usepackage{tikz}
\usetikzlibrary{backgrounds}
\usetikzlibrary{positioning} 
\usetikzlibrary{calc}
\usetikzlibrary{patterns} 
\usetikzlibrary{quotes} 
\usetikzlibrary{decorations.markings}
\usetikzlibrary{shapes.geometric}
\usetikzlibrary{fit}
\tikzset{
  fitting node/.style={
    inner sep=0pt,
    fill=none,
    draw=none,
    reset transform,
    fit={(\pgf@pathminx,\pgf@pathminy) (\pgf@pathmaxx,\pgf@pathmaxy)}
  },
  reset transform/.code={\pgftransformreset}
}
\usepackage{pgfplots}
\pgfplotsset{compat=1.16}
\usepgfplotslibrary{fillbetween}

\tikzset{
    database/.style={
        cylinder,
        aspect=0.5,
        draw,
        rotate=90,
        path picture={
            \draw (path picture bounding box.155) to[out=130,in=230] (path picture bounding box.25);
            \draw (path picture bounding box.180) to[out=130,in=230] (path picture bounding box.0);
        }
    }
}

\pgfkeys{/pgf/number format/.cd,fixed,precision=3}
\pgfplotsset{
    compat=1.3, 
    every axis/.append style={
        height=4cm,
        width=0.38\textwidth,
        xmin=0,
        xmax=10,
        ymin=0,
        ymax=100,
        xtick distance={3},
        ytick distance={30},
        minor xtick={1, 2, ..., 10},
        minor ytick={10, 20, ..., 100},
        grid=both,
        grid style={line width=.1pt, draw=gray!15},
        xtick pos=bottom,
        ytick pos=left,
    },
    yticklabel style={
        text width=0em,
        align=right
    },
}

\usepackage{siunitx}

\usepackage{caption}
\usepackage{subcaption}

\usepackage{lastpage}
\jmlrheading{XX}{2023}{1-\pageref{LastPage}}{2/23; Revised X/23}{X/23}{XX-XXXX}{Scheikl et al.}

\ShortHeadings{LapGym}{Scheikl et al.}
\firstpageno{1}

\begin{document}

\title{LapGym - An Open Source Framework for Reinforcement Learning in Robot-Assisted Laparoscopic Surgery}

\author{%
    \name Paul Maria Scheikl$^{1,\ 2}$ \email paul.scheikl@kit.edu \\
    \name Balázs Gyenes$^1$ \email balazs.gyenes@kit.edu \\
    \name Rayan Younis$^3$ \email rayan.younis@stud.uni-heidelberg.de \\
    \name Christoph Haas$^1$ \email christoph.haas@student.kit.edu \\
    \name Gerhard Neumann$^1$ \email gerhard.neumann@kit.edu \\
    \name Martin Wagner$^{3,\ 4 \ \dag}$ \email martin.wagner@med.uni-heidelberg.de \\
    \name Franziska Mathis-Ullrich$^{1, \ 2\ \dag}$ \email franziska.ullrich@kit.edu \\    
    \addr
    $^1$ Institute for Anthropomatics and Robotics (IAR),\\ Karlsruhe Institute of Technology (KIT),  76131 Karlsruhe, Germany\\
    $^2$ Health Robotics \& Automation (HERA), Department Artificial Intelligence in Biomedical Engineering (AIBE), Friedrich-Alexander-University Erlangen-Nürnberg (FAU), 91052 Erlangen, Germany\\
    $^3$ Department for General, Visceral and Transplantation Surgery,\\ Heidelberg University Hospital (UKHD),  69120 Heidelberg, Germany\\
    $^4$ Center for the tactile Internet with Human in the loop (CeTI),\\ Technical University Dresden (TUD), 01062 Dresden, Germany\\
    $^\dag$ These authors contributed equally.
}

\editor{TODO}

\maketitle

\begin{abstract}
Recent advances in reinforcement learning (RL) have increased the promise of introducing cognitive assistance and automation to robot-assisted laparoscopic surgery (RALS).
However, progress in algorithms and methods depends on the availability of standardized learning environments that represent skills relevant to RALS.
We present LapGym, a framework for building RL environments for RALS that models the challenges posed by surgical tasks, and \sofaenv, a diverse suite of $12$ environments.
Motivated by surgical training, these environments are organized into $4$ tracks: Spatial Reasoning, Deformable Object Manipulation \& Grasping, Dissection, and Thread Manipulation.
Each environment is highly parametrizable for increasing difficulty, resulting in a high performance ceiling for new algorithms.
We use Proximal Policy Optimization (PPO) to establish a baseline for model-free RL algorithms, investigating the effect of several environment parameters on task difficulty.
Finally, we show that many environments and parameter configurations reflect well-known, open problems in RL research, allowing researchers to continue exploring these fundamental problems in a surgical context.
We aim to provide a challenging, standard environment suite for further development of RL for RALS, ultimately helping to realize the full potential of cognitive surgical robotics.
LapGym is publicly accessible through GitHub (\url{https://github.com/ScheiklP/lap_gym}).
\end{abstract}

\newpage
\section{Introduction}
    \label{sec:introduction}
    \paragraph{Motivation}
        \Gls{rals} is nowadays gold-standard and has proven benefits in prostate surgery~\citep{soodRobotAssisted2014, stolzenburgRoboticAssisted2021}. 
        Because of its faster learning curve it became de facto standard of minimally invasive surgery for example in gynecology~\citep{jorgensenNationwideIntroduction2019} and advanced minimally invasive oncological visceral surgery~\citep{machadoTheRoleOf2021}.
        From a technical perspective, \gls{rl} has seen a rapid increase in research interest in the robotics community across a large range of applications such as legged locomotion and grasping~\citep{ibarzHowToTrain2021}.
        Progress on \gls{rl} in robotics depends on the availability of standardized learning environments that cover the required features and needs of the considered robotic application.
        However, the features required for learning skills that are relevant to \gls{rals} are not yet represented in a standardized suite of \gls{rl} environments.
        These features include collision-based interactions with deformable objects, topological changes introduced through surgical preparation, motion restricted by the \gls{rcm} in trocars for laparoscopic surgery, and simultaneous control of multiple instruments.
        In addition, images from the endoscopic camera are the primary source of information in surgical settings, but many existing environment suites do not support image observations.
        Recent works for automation in \gls{rals} propose algorithmic advances for solving surgical subtasks, but evaluate these algorithms in a) handcrafted simulation environments that are often not publicly available~\citep{shinAunotomousTissue2019, scheiklCooperativeAssistanceRobotic2021, heManipulatingConstrained2022, bourdillonIntegrationOfRL2022} or b) on expensive robotic hardware such as the \gls{dvrk}~\citep{tagliabueSoftTissue2020, poreLFD2021, chiuBimanualRegrasping2021, varierCollaborative2022}.
        This results in a high barrier for researchers to contribute to the field, and ultimately impedes vital algorithmic development.
        Furthermore, comparative validation of algorithms is difficult if not impossible.
        
        Recent \gls{rals} learning environments simulate a single, concrete task, and remain fixed during algorithm development and hyperparameter tuning.
        This results in a sparse feedback signal for researchers, as the tasks do not allow for fine-grained comparisons across various difficulty levels.
        Thus, incremental improvement may become hard to detect when even good algorithms are unlikely to solve the task at all.
        For this reason, we argue that an ideal environment should represent a class of \gls{rl} problems where the difficulty can be adjusted through several parameters.
        
        A large barrier for \gls{rl} researchers from other domains exists to additionally validate their approaches for the surgical domain, as easy-to-use, surgically relevant \gls{rl} environments do not exist.
        We aim to provide the scientific community with easy-to-use, surgically relevant \gls{rl} environments to enable research seeking to improve over a whole suite of environments, allow for benchmarking methodologies, and further advance the field as a whole.
    
    \paragraph{Contributions}
        Here, we propose LapGym, an open-source framework for \gls{rl} in \gls{rals} that includes learning environments for challenging surgical tasks (\sofaenv), utilities to implement new tasks and environments (\sofagodot and \sofatemplates), and baseline \gls{rl} experiments (\sofazoo).
        An overview of LapGym is shown in \autoref{fig:LapGym}.
        Path planning, human control, and sensor simulation functions allow generation of robotic and expert trajectories based on multi-modal sensor data for imitation learning. 
        
        \sofaenv includes a set of $12$ learning environments grouped into $4$ surgical training tracks that cover a broad spectrum of skills required for \gls{rals}.
        
        The environments:
        \begin{itemize}
            \item Employ \gls{sofa} as the numerical physics simulation for \gls{fem} simulation of deformable objects.
            \item Are highly parametrizable, allowing for a gradual increase of their difficulties to facilitate incremental algorithm development and hyperparameter tuning.
            \item Are provided fully open-source and publicly accessible through GitHub under the MIT license.
        \end{itemize}
        The \gls{rl} experiments in \sofazoo establish a baseline using \gls{ppo}~\citep{schulmanProximalPolicyOptimization2017} to show initial performance of a well established algorithm across different configurations of the environments.
        
        Open problems in \gls{rl} and \gls{rals} research are represented throughout the environment suite, making it a suitable testbed for further research into these problems.
        In summary, we provide the community with a challenging benchmark that makes it possible to compare different \gls{rl} approaches against each other, and advance the state of the art for \gls{rl} in \gls{rals}.

    \input{figures/overview.tex}

\section{Related Work}
    \label{sec:relatedwork}
    Popular reinforcement learning suites cover a multitude of applications: game environments~\citep{brockmanOpenaiGym2016}, closed-loop control tasks of articulated rigid objects~\citep{tassaDeepmindControl2018, yuMetaWorld2020}, control of soft robotic actuators~\citep{menagerSofaGym2020}, robot-specific learning environments~\citep{varierAMBFRL2022, richterDVRL2019, xuSurRol2021, tagliabueSoftTissue2020}, musculoskeletal motor control~\citep{caggianoMyoSuite2022}, and toy environments with simplified deformations~\citep{laezzaReform2021, linSoftGym2020}.
    
    \citet{menagerSofaGym2020} propose SofaGym, which targets \gls{rl} for the control of soft robotic actuators.
    While SofaGym also utilizes \gls{sofa} as its physics simulator, the software architecture and implemented features are tailored to control of soft robotics, making it impractical for extension to \gls{rals}.
    Creation of new learning environments is possible, but requires in-depth knowledge about the architecture and usage of \gls{sofa}, proving to be its main limitation.
    
    In robot-assisted surgery, \citet{richterDVRL2019} propose dVRL, a \gls{rl} framework to control a \gls{dvrk} robot in reaching tasks.
    These tasks include moving the \gls{dvrk}'s end-effector to desired positions to pick up and place objects, or for suction of liquids.
    The simulation employs CoppeliaSim (formerly known as VREP, \cite{rohmerVREP2013}) and does not support deformable objects.
    In addition, interaction between objects is static, such that grasping is implemented by attaching one physical object to another, without actual simulation of grasping through collision and friction.
    
    \citet{xuSurRol2021} simulate rigid object interaction by employing the Bullet Physics library in their simulation platform, SurRoL~\citep{coumansBullet2015}.
    SurRoL, like dVRL, focuses on the \gls{dvrk} as a robotic platform and implements several rigid body tasks such as picking up needles and performing peg transfer, which present common tasks in surgical training.
    
    \citet{varierAMBFRL2022} propose AMBF-RL, which implements an \gls{rl} interface for the robotic simulator AMBF~\citep{munawarAMBF2019}.
    \citeauthor{varierAMBFRL2022} evaluate their framework on a reaching task, where a \gls{dvrk} robot is controlled to reach a desired point in Cartesian space, but do not provide any additional learning environments.
    AMBF's deformable object simulation relies on the Bullet Physics library~\citep{coumansBullet2015} for simulation using \gls{pbd}~\citep{mullerPositionBased2007}.
    Therefore, while AMBF technically supports simulation of deformable objects, none of the implemented simulation scenes actually contain deformable objects beyond simple beam structures such as a suturing thread.
    This is likely due to the complexity of tuning and the numerical instabilities of deformable object simulation in Bullet.
    Furthermore, AMBF-RL depends on the Melodic release of the Robot Operating System (ROS)~\citep{quigleyROS2009} and Ubuntu 18.04, most likely for interoperability with the \gls{dvrk}.
    While ROS can speed up development of robotic systems, it poses a heavy software liability, and is currently being phased out in favor of ROS 2 with limited backwards compatibility.
    
    UnityFlexML \citep{tagliabueSoftTissue2020} was the first \gls{rl} framework that supports simulation of deformable objects.
    \citeauthor{tagliabueSoftTissue2020} implement the task of tissue retraction in Unity, utilizing Nvidia Flex as a physics engine.
    However, Nvidia Flex has been discontinued, and Linux support for Unity is still limited, making it difficult to create reliable \gls{rl} environments.

    None of the presented works support image observations, which are a crucial feature for \gls{rals}, as state observations are difficult to define and extract from endoscopic images in a real world setting.

    This work proposes \sofaenv, a \gls{rl} framework built on the \gls{sofa} simulation framework.
    \sofaenv has few dependencies other than \gls{sofa}, mainly popular Python libraries such as numpy, pyglet, and Open3D.
    Using pyglet and Open3D, \sofaenv supports various image based observation types such as RGB images, depth images, point clouds, and semantically segmented images.
    Using \gls{sofa}, \sofaenv supports simulation of rigid and deformable objects, topological changes to simulate surgical preparation of deformable objects, contact rich interaction between objects, and fast \gls{fem} simulation, even allowing for interactive control by a human operator.

\section{Technical Requirements for Robot-Assisted Laparoscopic Surgery}
    \label{sec:technical}
    The following are characteristic challenges in laparoscopic surgery, that are essential for learning useful skills for \gls{rals}:
    
    \paragraph{Image observations}
        Most works on \gls{rl} for \gls{rals} hand-craft state observations, which are comprised of a set of features that are extracted through complex and labor-intensive image processing pipelines, such as describing the deformed state of organs and tissues.
        However, in clinical reality, these complex pipelines are rarely feasible.
        During surgery, the only readily available source of information is an endoscope that captures image observations of the surgical site.
        In most cases, human surgeons perform laparoscopic surgery based on visual information on a 2D screen~\citep{zundelSuggestion2019}.
        This indicates that human surgeons have a mental model for spatial reasoning that they built through extensive training.
        However, in \gls{rals}, the trend is to provide human surgeons with depth information through stereo cameras~\citep{tsudaSages2015}.
        Image observations (RGB and depth), which also more closely represent the clinical reality, present a core component of \sofaenv and can be generated from simulated endoscopic cameras.
        
    \paragraph{Topological changes}
        Topological changes to deformable objects as a result of surgical preparation (\eg, cutting) is an essential part of the vast majority of surgical interventions.
        \sofaenv supports cutting of deformable objects using the SofaCarving plugin, which removes elements from the object topology based on detected collisions with a cutting instrument.
        SofaCarving is utilized to simulate electrocautery hooks and laparoscopic scissors in \sofaenv.

\begin{figure}[tbh]
    \centering
    \begin{tikzpicture}
        \node[anchor=center] at (0, 0) {\includegraphics[width=0.4\textwidth]{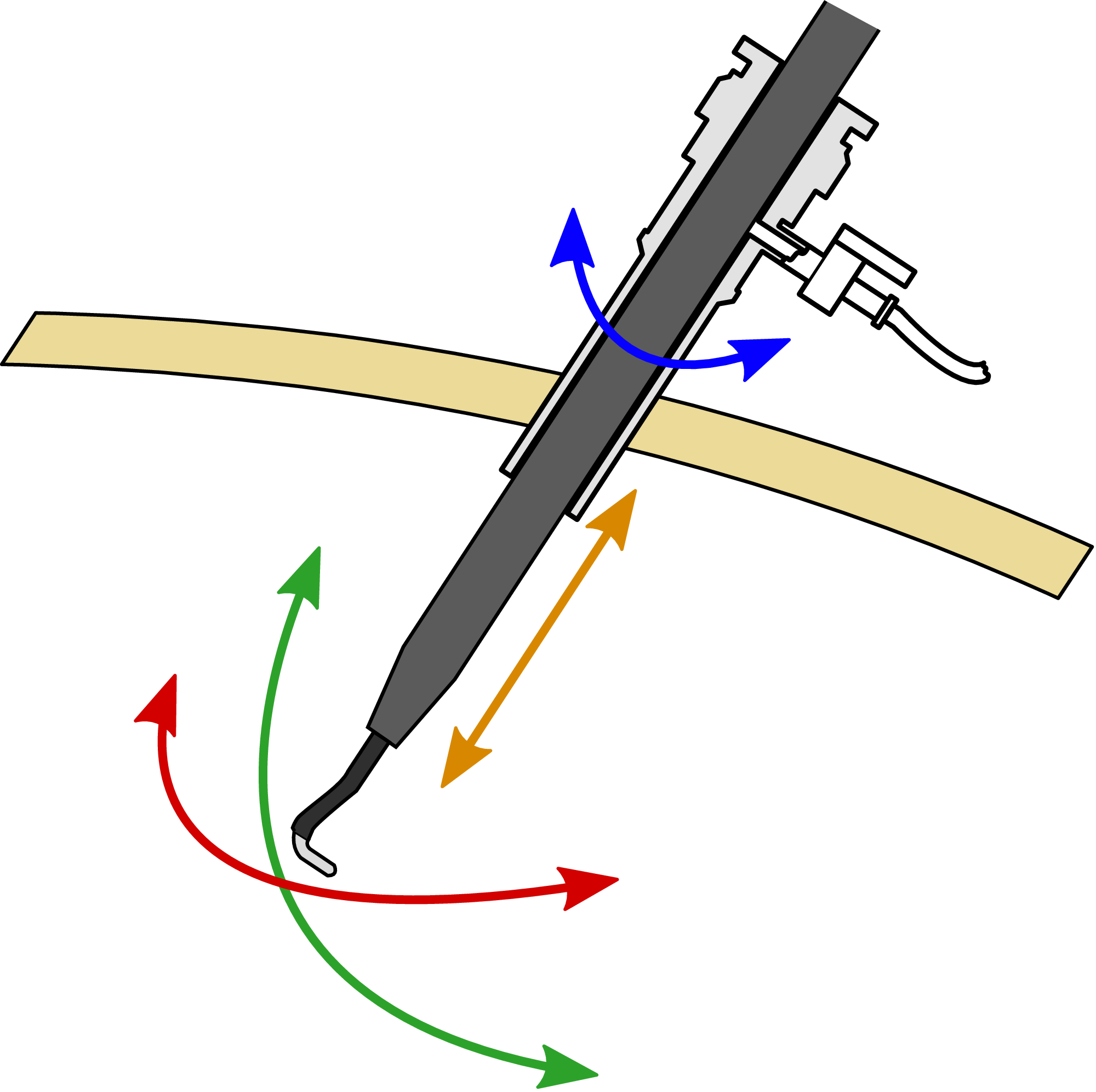}};
        \node[anchor=east] at (-2, -0.5) {\bf pan};
        \node[anchor=south west] at (0.3, -3.1) {\bf tilt};
        \node[anchor=east] at (0, 1.75) {\bf spin};
        \node[anchor=west] at (0, -0.7) {\bf depth};
        \node[anchor=south] at (-2.5, 1.2) {\begin{tabular}{c} \bf abdominal \\ \bf wall \end{tabular}};
        \node[anchor=west] (rcm) at (2.2, 1.5) {\bf RCM};
        \draw[fill=black, draw=white, ultra thick] (0.28, 0.8) circle (0.15);
        \draw[dashed, ultra thick, gray] (rcm.south) -- (0.28, 0.8);
    \end{tikzpicture}
    \caption{
        Laparoscopic surgery is performed with rod-shaped tools that are inserted into the abdominal cavity through small incisions in the abdominal wall.
        The pose of the instrument tip is constrained to pivotized motion around the \gls{rcm}, such that the degrees of freedom are reduced to $4$ independent values the represent $3$ rotations (tilt, pan, spin) and one translation (depth).
    }
    \label{fig:pivotization}
\end{figure}

    \paragraph{Pivotized motion}
        Laparoscopic instruments feature fewer than $6$ degrees of freedom due to the \gls{rcm} constraint imposed by a trocar at which they pass the patient's abdominal wall~\citep{nisarRCM2017}, as illustrated in \autoref{fig:pivotization}.
        An \gls{rl} agent must learn to control the Cartesian position of the end-effector, while satisfying this constraint.
        We define the controllable degrees of freedom as Tilt, Pan, Spin, and Depth (TPSD).
        The first three (\ie pan, tilt, spin) define rotations in relation to the coordinate system defined in the \gls{rcm}, while depth is defined as translation along the longitudinal axis of the instrument shaft from the \gls{rcm}.
       
        An affine transformation describes the forward kinematics of a laparoscopic instrument.
        The transformation that maps a TPSD state to the 6D pose of a laparoscopic instrument inserted at a \gls{rcm} in world coordinates is described by a homogeneous matrix $T$:
        \begin{equation}
            T = T_t(\text{RCM position}) * T_{xyz}(\text{RCM orientation}) * T_{xyz}(\text{tilt, pan, spin}) * T_t([0, 0, depth]) 
        \end{equation}
        with homogeneous matrices $T_t$ to represent a translation and $T_{xyz}$ to define a rotation with XYZ Euler angles.
        
        Several laparoscopic instruments such as endoscopes, electrocautery hooks, and different graspers are implemented in \sofatemplates.
        These instruments are controlled in TPSD space and can be constrained with limits in both Cartesian and TPSD workspaces.
            
\section{Software Framework}
    \label{sec:framework}
    LapGym builds on the interactive \gls{fem} simulation \gls{sofa} as a physics simulation to create \gls{rl} environments for \gls{rals}.
    The framework further provides utilities that simplify creation of \gls{rl} environments for existing \gls{sofa} simulation scenes, as well as creating new \gls{rals} scenes without in-depth knowledge of \gls{sofa}.
    
    \subsection{SOFA}
        The simulation framework \gls{sofa} is a widely used tool in the field of medical simulations.
        Its advanced physics-based modeling capabilities make it an ideal platform for developing and testing control algorithms for robotic surgical systems.
        In contrast to other \gls{fem} simulators, \gls{sofa} is capable of running simulations at interactive speeds, which is crucial for methods that require large amounts of data, such as \gls{rl}.
        This characteristic not only enables rapid iteration and optimization of learning algorithms, it also enables realistic human interaction, which is relevant for recording (expert) demonstrations and validating human-robot collaboration.
        \gls{sofa} is an active open source project that is continuously increasing in popularity, attracting new contributors and accruing new features such as dynamic mesh refinement, control of magnetic continuum robots~\citep{dreyfusMCR2022}, and synthetic data generation from numerical simulations for training machine learning models~\citep{mimesisDeepPhysX2022, linkerhaegnerGrounding2023}.

    \subsection{SOFA for Reinforcement Learning}
        \gls{sofa} simulation scenes are defined by a scene graph (directed acyclic graph) of components such as numerical solvers and elements that model object behavior (\eg deformation, visual, collision).
        The scene graph is usually defined in a Python script using the popular SofaPython3 plugin, which exposes \gls{sofa} functions and types through Python bindings.
        Such a Python script must implement a \code{createScene()} function, which receives the root node of the scene graph as input and adds desired components to the graph.
        The \code{runSofa} binary uses this script to load and run the simulation, and opens a GUI for the user to observe the rendered scene.
        Instead, the SofaEnv environment base class of \sofaenv uses SofaPython3 to instantiate a \gls{sofa} simulation programmatically based on the \code{createScene()} function.
        References to objects such as laparoscopic instruments, cameras, or the mechanical state of a deformable object are passed to the environment after creation, allowing the simulation state to be read and manipulated by Python code.
        Each environment's step function uses \gls{sofa} object handles to update positions, forces, velocities, \etc and advance the simulation.
        Likewise, the environment can read data from the \gls{sofa} simulation to calculate rewards and retrieve observations.

        The stability of the \gls{sofa} simulation depends on the length of the time step $\Delta T_s$.
        Lower $\Delta T_s$ result in a more accurate and stable simulation.
        However, the difference between consecutive observations may become very small (\eg, \SI{0.001}{\second}), which is undesirable in \gls{rl}, as this makes it more difficult to learn how actions affect the observed environment state.
        A common solution is frame skipping~\citep{mnihPlaying2013}, where the same action is applied $n$ times to the environment, with intermediate observations discarded.
        SofaEnv environments use frame skipping to decouple the time scale of the simulation from the time scale of \gls{rl}.

        The SofaEnv base class implements the widely-used \textit{Gym} interface from OpenAI~\citep{brockmanOpenaiGym2016}, and adheres to Gym's specification for observation and action spaces.
        SofaEnv environments are thus out-of-the-box compatible with well-known \gls{rl} environments such as StableBaselines3~\citep{raffinStableBaselines3ReliableReinforcement2021} and RLlib~\citep{liangRLlib2018}.

    \subsection{Implementation of New Environments}
        The main limitation of \gls{sofa} is the required specific knowledge about \gls{fem} in general and \gls{sofa}'s software architecture in particular to achieve the desired simulation behavior.
        This work addresses this limitation by providing the user with a set of high level templates that combine all required components with sane default values.
        These templates are provided through \sofatemplates and aim to simplify the process of writing \gls{sofa} simulation scenes, also independent of creating \gls{rl} environments through \sofaenv.
        Complex simulation subgraphs in \gls{sofa} such as partially rigidifying a deformable volumetric mesh are abstracted into utility functions that take simple arguments such as the indices of the object that should be rigidified and create the subgraph automatically.
        For example, when adding a deformable object to a simulation scene, the user can instantiate a \code{DeformableObject} class from the \sofatemplates, passing a volumetric mesh for \gls{fem} simulation and optional surface meshes for visual and collision models.
        This feature obviates the need for manually defining the numerical solvers, constraint corrections, force fields to describe the mechanical behavior, collision model behavior, and the correct mappings between the components.
        For users that want more fine-grained control over the behavior, the templates allow passing functions that overwrite default behavior, e.g. custom collision models or numerical solvers.
        Existing \gls{sofa} scenes can be converted to \gls{rl} environments simply by defining how an agent's action should affect the objects in the simulation (\eg by applying the action as a force on a robotic joint), and by defining a reward function.
        Most code in the \code{createScene()} function requires no additional modifications.
        
\begin{figure}[tbh]
            \centering
            \begin{subfigure}[b]{0.4\textwidth}
                \centering
                \includegraphics[height=4cm]{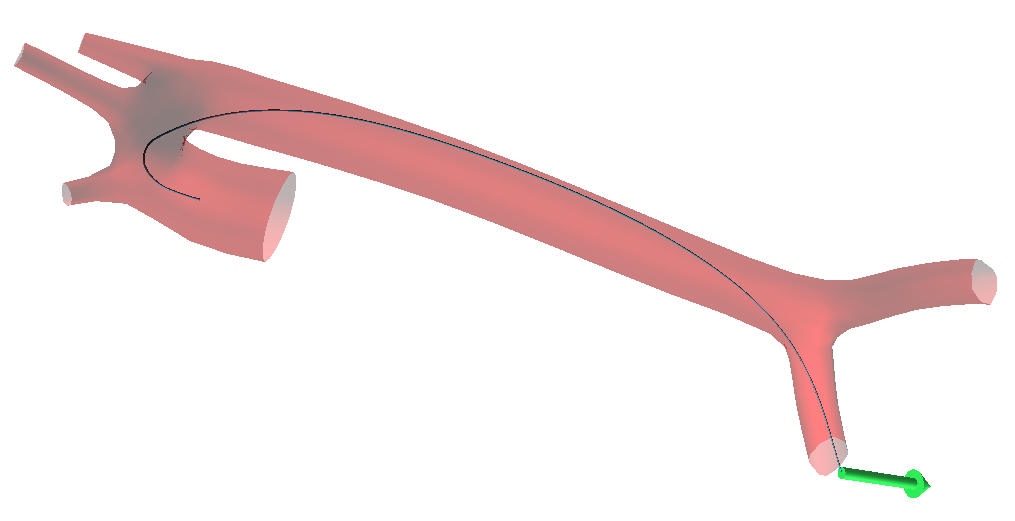}
                \caption{}
            \end{subfigure}
            \hfill
            \begin{subfigure}[b]{0.4\textwidth}
                \centering
                \includegraphics[height=4cm]{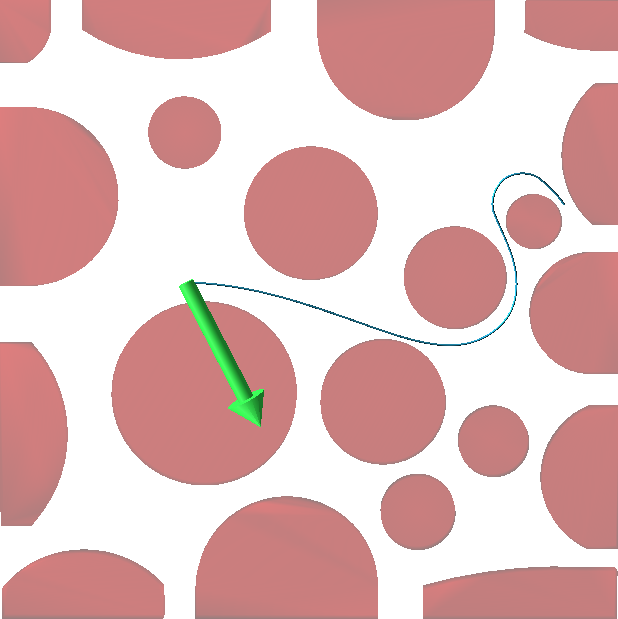}
                \caption{}
            \end{subfigure}
            \caption{Magnetic continuum robot scene from \citet{dreyfusMCR2022} for navigation in (a) an aortic arch model and (b) a planar toy environment. 
            Images by courtesy of~\citeauthor{dreyfusMCR2022}.}
            \label{fig:mcr}
        \end{figure}

        The aim of these tools is to provide both experienced \gls{sofa} users and newcomers with an easy-to-use framework for \gls{rl}.
        To demonstrate \sofaenv's ease of use, we port the existing \gls{sofa} simulation from \citet{dreyfusMCR2022} and define a \gls{rl} environment for its task.
        The scene contains a magnetic continuum robot that is controlled through an external force field in order to navigate in a model of (a) an aortic arch and (b) a 2D toy scene, as shown in~\autoref{fig:mcr}.
        We decided on this scene as it is comprised of complex simulation components, and to show that \sofaenv can be used beyond the scope of \gls{rals}.
        The scene description underwent only minor changes to comply with the most recent \gls{sofa} version.
        We provide a \gls{rl} environment for this scene in the source code, but do not consider this environment for the following experiments, as it is not in the scope of \gls{rals}.
        We also include the code to reproduce the learning environment of \citet{scheiklSimToReal2023}, and extend the environment of \citet{scheiklCooperativeAssistanceRobotic2021}.
        
    \subsection{Interactive Scene Creation}
        \label{subsec:sofagodot}
\begin{figure}[tbh]
            \centering
            \includegraphics[width=0.99\textwidth,keepaspectratio]{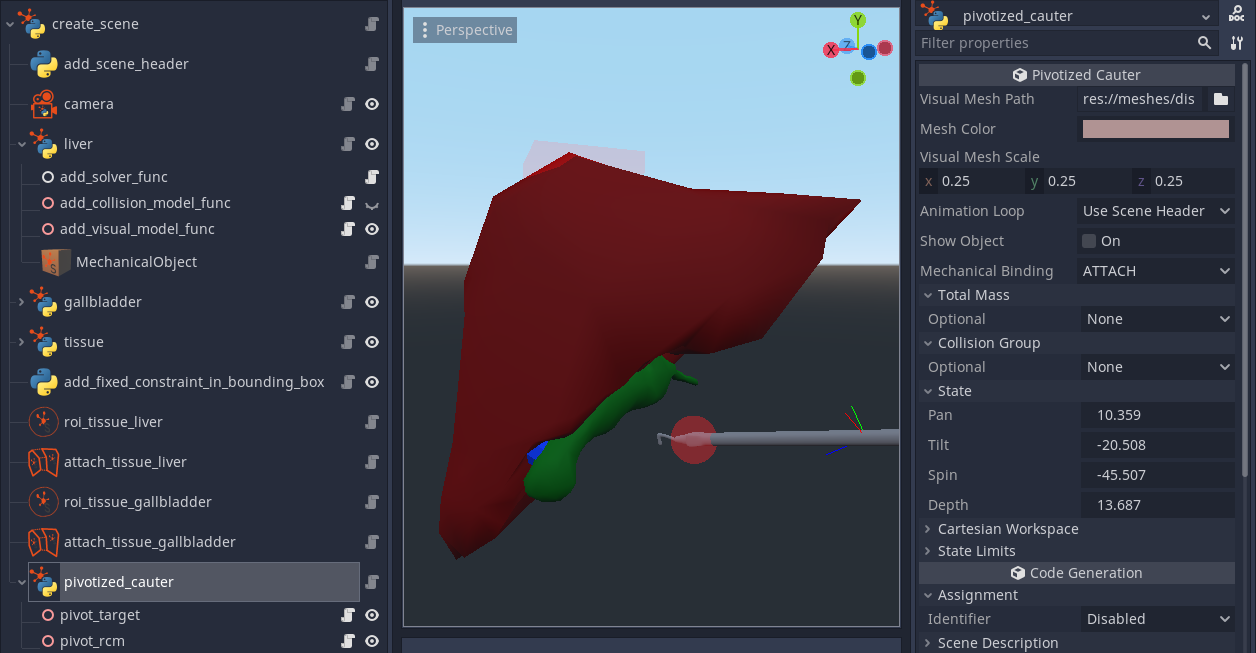}
            \caption{
                Graphical user interface of the Godot game engine displaying the prototype scene of a gallblader removal (Cholecystectomy) built with the \sofagodot plugin.
                Left: Scene graph comprised of specialized \sofagodot nodes corresponding to components from \gls{sofa} and \sofaenv.
                Center: 3D scene view showing visualizations of deformable objects, \eg liver and gallblader, as well as a pivotized instrument used for resection.
                Right: Display of editable properties from the currently selected node.
            }
            \label{fig:SofaGodot}
        \end{figure}

        Despite the advantages of \sofaenv and the included \sofatemplates, correctly placing objects in the scene and adding motion constraints, such as attachments, is still an arduous trial-and-error process.
        The default \textit{runSofa} binary does not allow the user to edit and visualize the scene simultaneously.
        To simplify this process, we introduce \sofagodot, a plugin for the popular free and open-source game engine Godot~\citep{linietskyGodot2014}.
        As shown in \autoref{fig:SofaGodot}, \sofagodot allows interactively building \gls{sofa} scenes through a \gls{gui} with a 3D scene view.
        \sofagodot features compatibility with SofaPython3 and \sofatemplates to further ease prototyping and development of \gls{rl} environments.
        The plugin provides a set of specialized Godot nodes, each of which implements a \gls{sofa} or \sofatemplates component with corresponding 3D visualization. 
        The parameters of a component are displayed as node properties and can be changed by the user through Godot's \gls{gui}, where the effects are visible instantly.
        \sofagodot automatically translates a Godot scene into a \gls{sofa} \code{createScene()} function in Python.
        The scene graph is traversed in a depth-first manner and every node visited can add Python statements to the body of the \code{createScene()} function.
        The arguments of these Python statements are derived from the properties of the respective node.
        This concept enables highly flexible wrapping of components, \ie Python statements, as Godot nodes.
    
    \subsection{Human Control and Path Planning}
        All environments include Python scripts that allow to control the instruments with an Xbox One controller (Microsoft Corp., USA) and to collect expert trajectories.
        The collected data can be used for imitation learning methods, or to establish a human baseline to compare learned agent behavior.
        The values recorded during a trajectory can be flexibly customized through a callback mechanism.
        We also provide an environment wrapper that is able to perform collision free motion planning with rapidly-exploring random trees (RRT)~\citep{lavalleRRT1998} in Cartesian and TPSD space to automatically generate trajectories.

\section{Learning Environments}
    \label{sec:envs}
    We present a suite of $12$ learning environments, which together provide a set of challenging tasks that span the skills required for \gls{rals}.
    On the basis of surgical experience in learning and teaching \gls{rals} as well as discussions with other surgical experts, the environments are sorted into $4$ surgical training tracks:
    1) Spatial Reasoning,
    2) Deformable Object Manipulation and Grasping,
    3) Dissection, and
    4) Thread Manipulation.
    Each of these tracks focuses on specific challenges in skill learning for laparoscopic surgery.
    Many of the environments were adapted from common surgical training tasks, \eg from the \gls{fls} program~\citep{petersFLS2004}, established surgical training simulators such as Simbionix (Surgical Science, Sweden) and SimNow (Intuitive Surgical, USA).
    This enables meaningful comparison between trained policies and human surgeons.
    
\begin{figure}[tbh]
            \centering
            \begin{subfigure}[b]{0.22\textwidth}
                \centering
                \includegraphics[width=\textwidth]{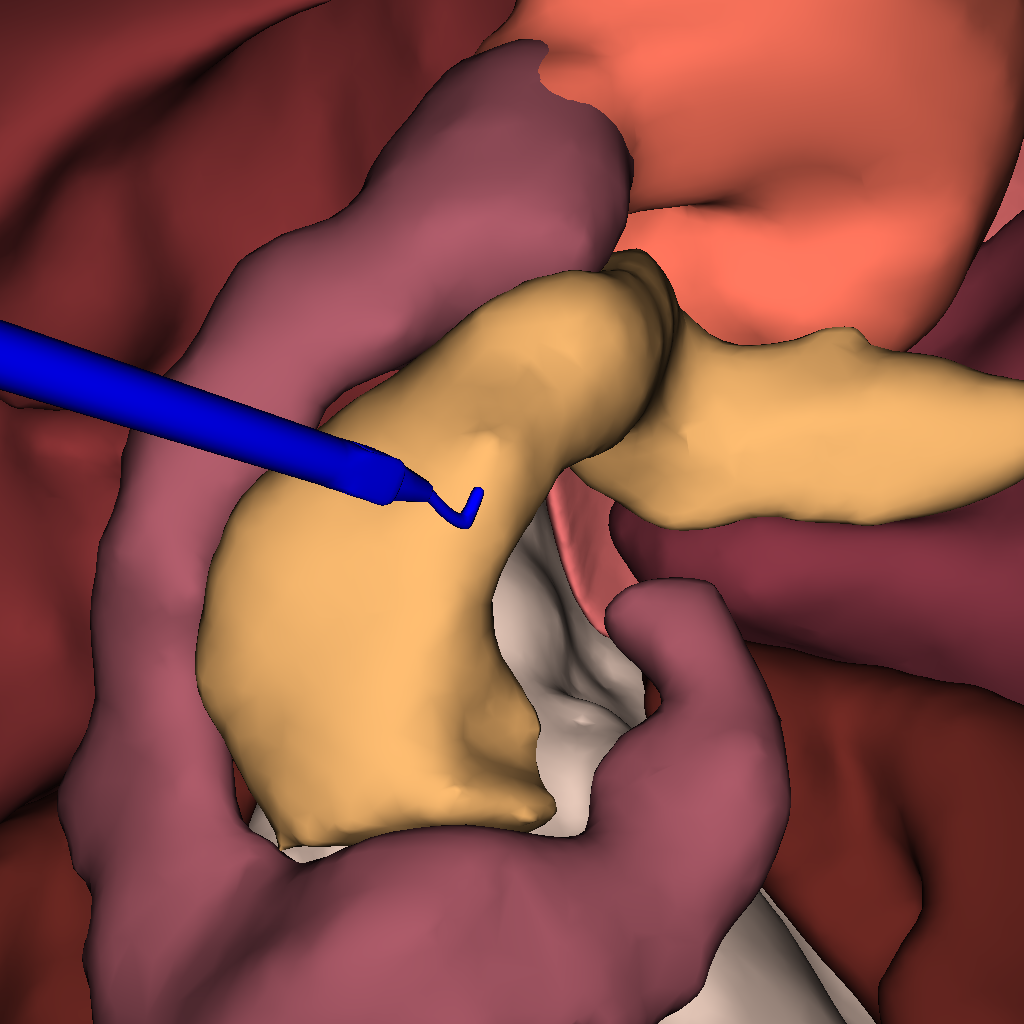}
                \caption{RGB}
            \end{subfigure}
            \hfill
            \begin{subfigure}[b]{0.22\textwidth}
                \centering
                \includegraphics[width=\textwidth]{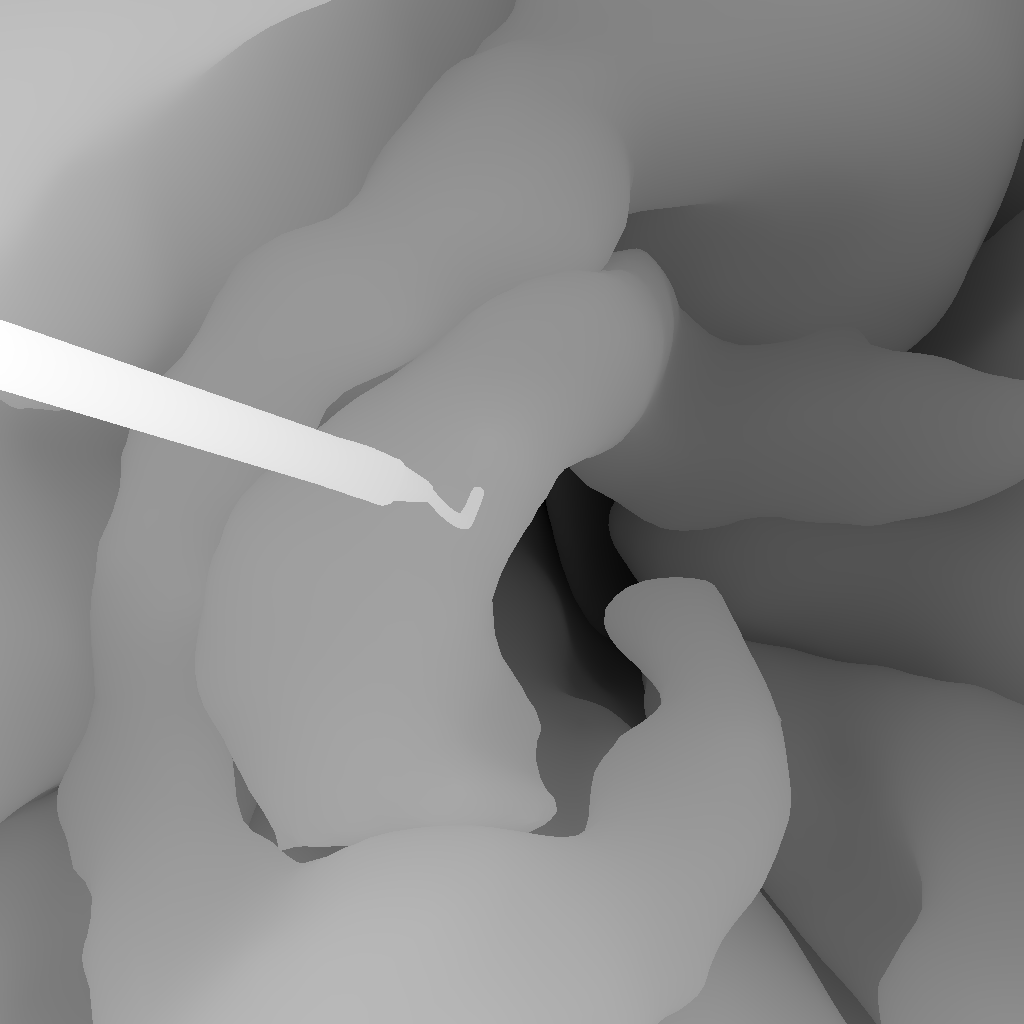}
                \caption{Depth}
            \end{subfigure}
            \hfill
            \begin{subfigure}[b]{0.22\textwidth}
                \centering
                \includegraphics[width=\textwidth]{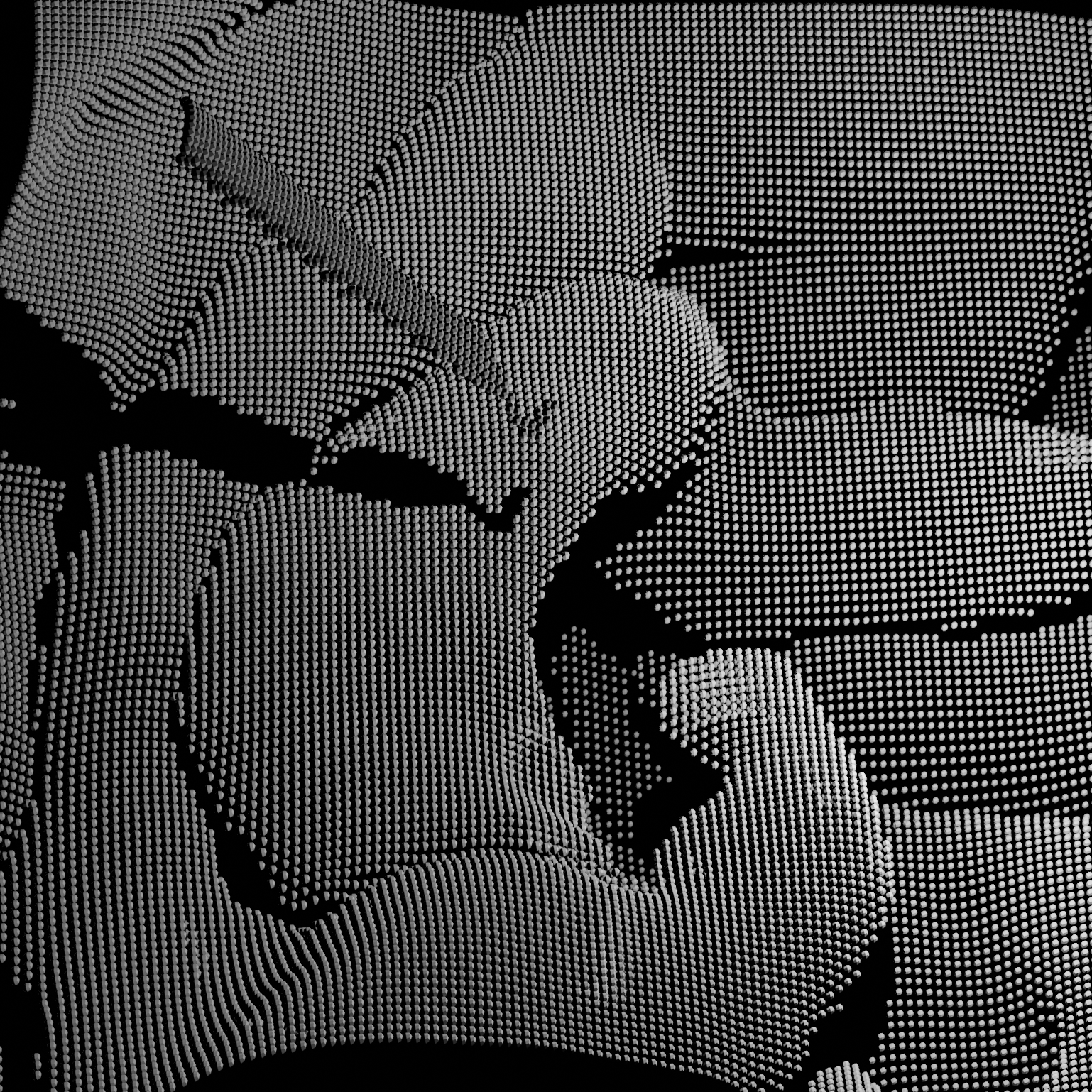}
                \caption{Point Clouds}
            \end{subfigure}
            \hfill
            \begin{subfigure}[b]{0.22\textwidth}
                \centering
                \includegraphics[width=\textwidth]{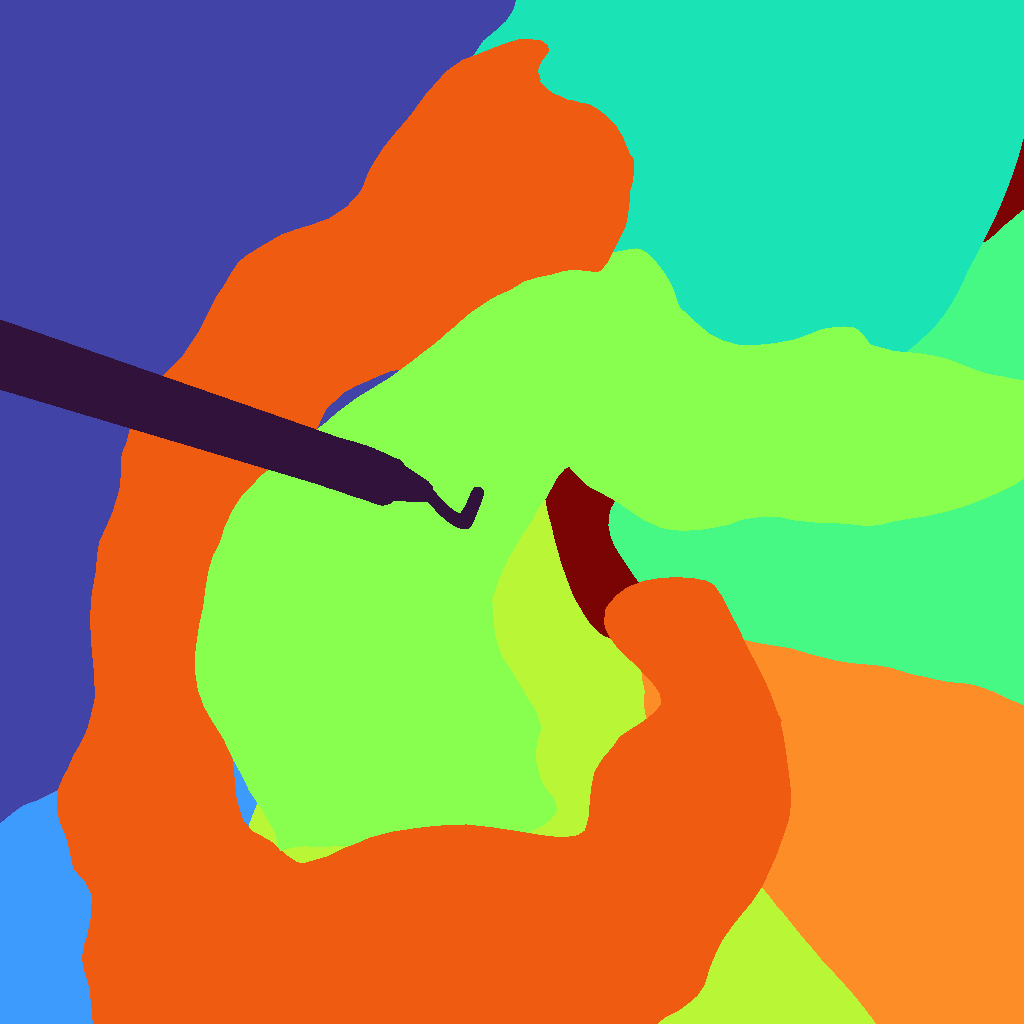}
                \caption{Semantic Seg.}
            \end{subfigure}
            \caption{Supported image observation types, visualized for SearchForPointEnv. (a)~RGB and  (b)~depth images are generated from OpenGL buffers, while (c)~point clouds and (d)~semantic segmentation images are generated through the Open3D library.}
            \label{fig:observationtypes}
        \end{figure}

    \paragraph{Observations and Actions}
        All environments define RGB, RGBD, and hand-crafted state observation spaces and allow for easy adaptation \eg to multi-modal observations.
        Wrappers provided by the \sofaenv utilities can be used to extend the observations to point clouds and semantically segmented images from one or more cameras (see \autoref{fig:observationtypes}).
        All environments support both continuous and discrete actions.
        
    \paragraph{Reward Functions}
        All environments define a hand-crafted set of reward features $\bm{\psi}$, but all environments can easily be subclassed to customize the reward function arbitrarily.
        By default, the total reward $r_t = \Sigma_i w_i\psi_i$ is a weighted sum of these features.
        We attempt to set reasonable defaults for these weights $\bm{w}$, but they are overridable by the user.
        Concerning safety, negative rewards are implemented to punish safety relevant features such as undesired collisions or workspace violations. 
        A description of the environment specific reward functions and their default weights can be found in Appendix A, Table A.
        
    \subsection{Spatial Reasoning Track}
        The spatial reasoning track mainly incorporates the challenges of estimating depth from 2D visual observations and navigating under the motion constraints introduced by the \gls{rcm}.
        These challenges are addressed in each environment of this track and get progressively more difficult, adding additional kinematic constraints that are familiar to laparoscopic surgeons.
        
        \begin{figure}[tbh]
            \centering
            \begin{subfigure}[b]{0.35\textwidth}
                \centering
                \includegraphics[width=\textwidth]{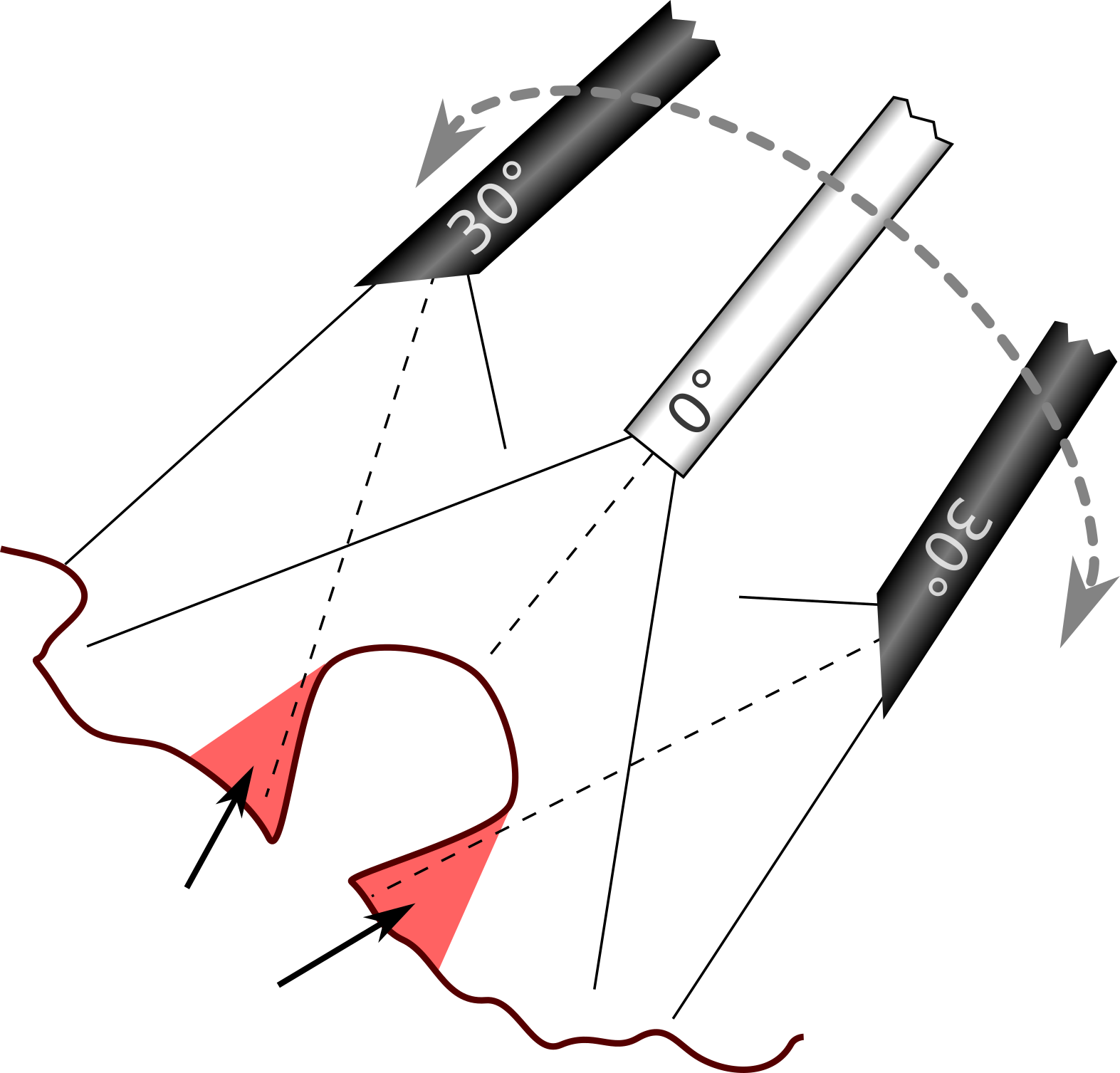}
                \caption{}
                \label{fig:undercut}
            \end{subfigure}\hfill
            \begin{subfigure}[b]{0.6\textwidth}
                \centering
                \begin{tikzpicture}
                    \node[anchor=center] at (0, 0) {\includegraphics[width=\textwidth]{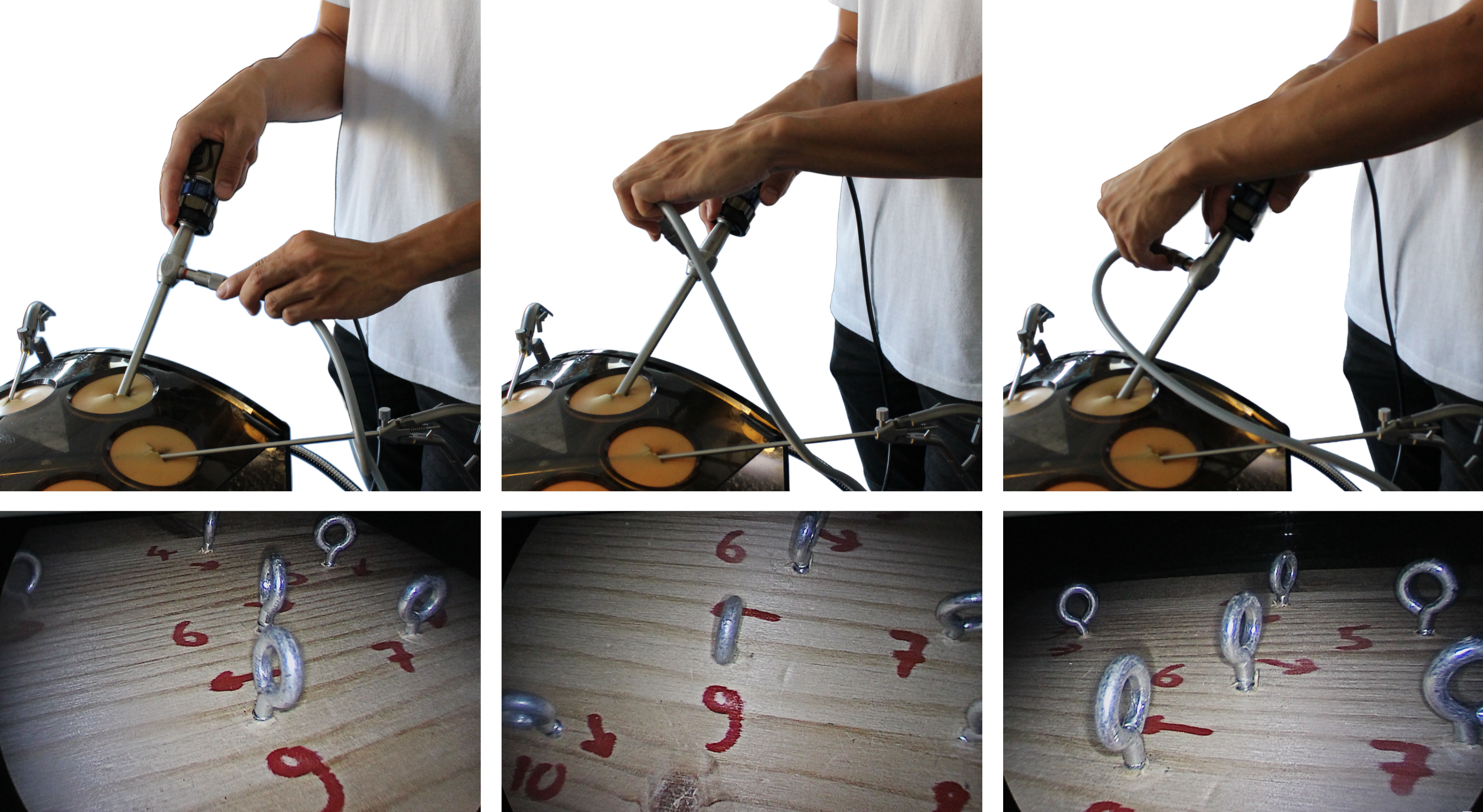}};
                    \draw[-stealth, ultra thick, red] (-3.6, 2) -- (-2.6, 2);
                    \draw[-stealth, ultra thick, red] (-2.2, 0.4) -- (-3.2, 0.4);
                \end{tikzpicture}
                
                \caption{}
                \label{fig:stableHorizon}
            \end{subfigure}
            \caption{
                (a) Oblique-viewing endoscopes ($30^{\circ}$) can obtain views otherwise occluded (marked in red) with $0^{\circ}$-endoscopes by rotating an angled optic relative to the image sensor.
                (b) To generate different views of the scene, the surgeon pans the camera head while rotating the optic relative to the camera head, thus looking \textit{from the right side} or \textit{from the left side} while keeping a stable horizon of the image relative to a global frame of reference.
            }
            \label{fig:oblique}
        \end{figure}

        \begin{tabularx}{\textwidth}{lX}
            \raisebox{-\totalheight}{\includegraphics[width=2.5cm]{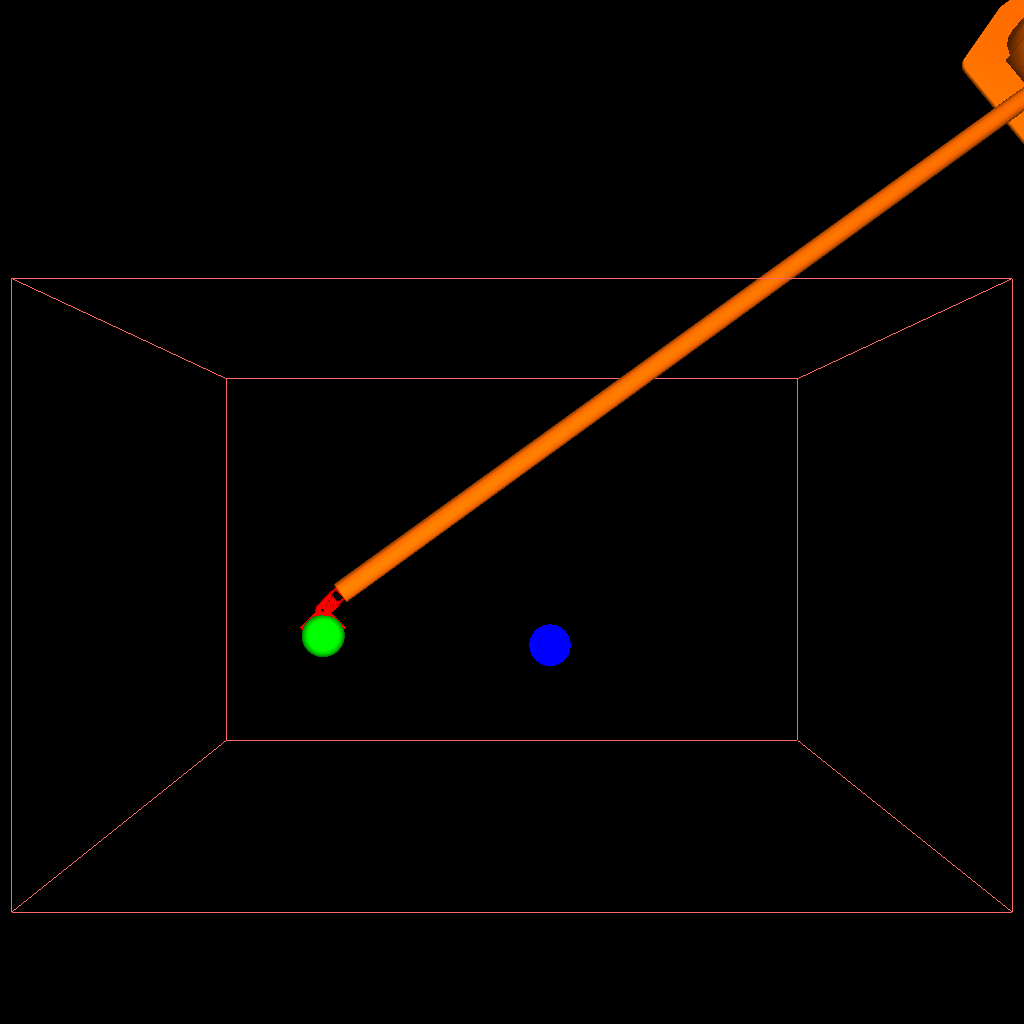}} &
            \hypertarget{env:reach}{\textbf{ReachEnv}} contains a surgical robotic end-effector with a sphere mounted to its end.
            The goal is to reach a target position, visualized by a colored sphere, by controlling the end-effector directly in Cartesian space.
            The task is finished when the distance between end-effector and target is lower than a defined threshold.\\
            \raisebox{-\totalheight}{\includegraphics[width=2.5cm]{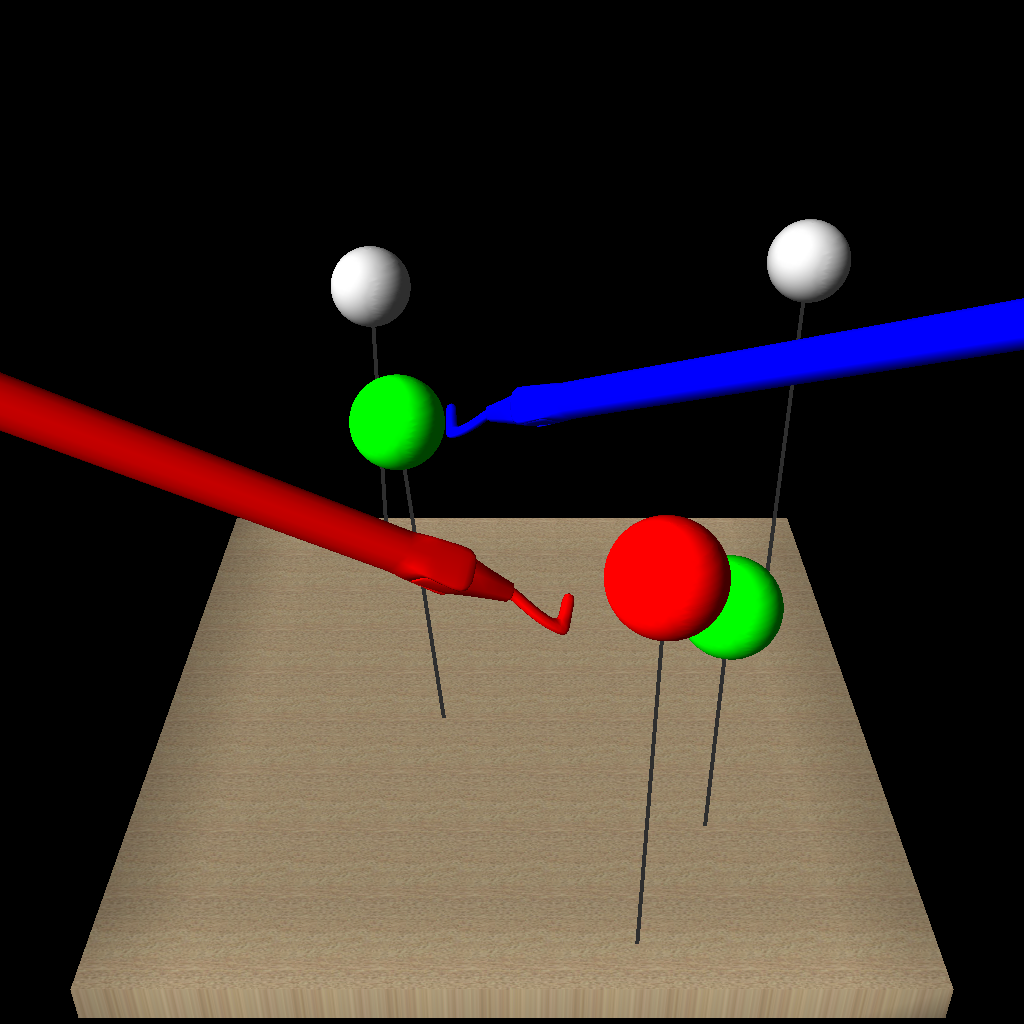}} &
            \hypertarget{env:deflect}{\textbf{DeflectSpheresEnv}} consists of a flat board with $N$ spheres mounted to the board on flexible stalks.
            Two electrocautery hooks, one blue and one red, are controlled in TPSD space.
            The goal is to collide with and deflect the highlighted active sphere using the instrument of matching color, which marks it as complete and changes its color to green.
            When $M$ spheres have been deflected correctly, the task is completed.\\
            \raisebox{-\totalheight}{\includegraphics[width=2.5cm]{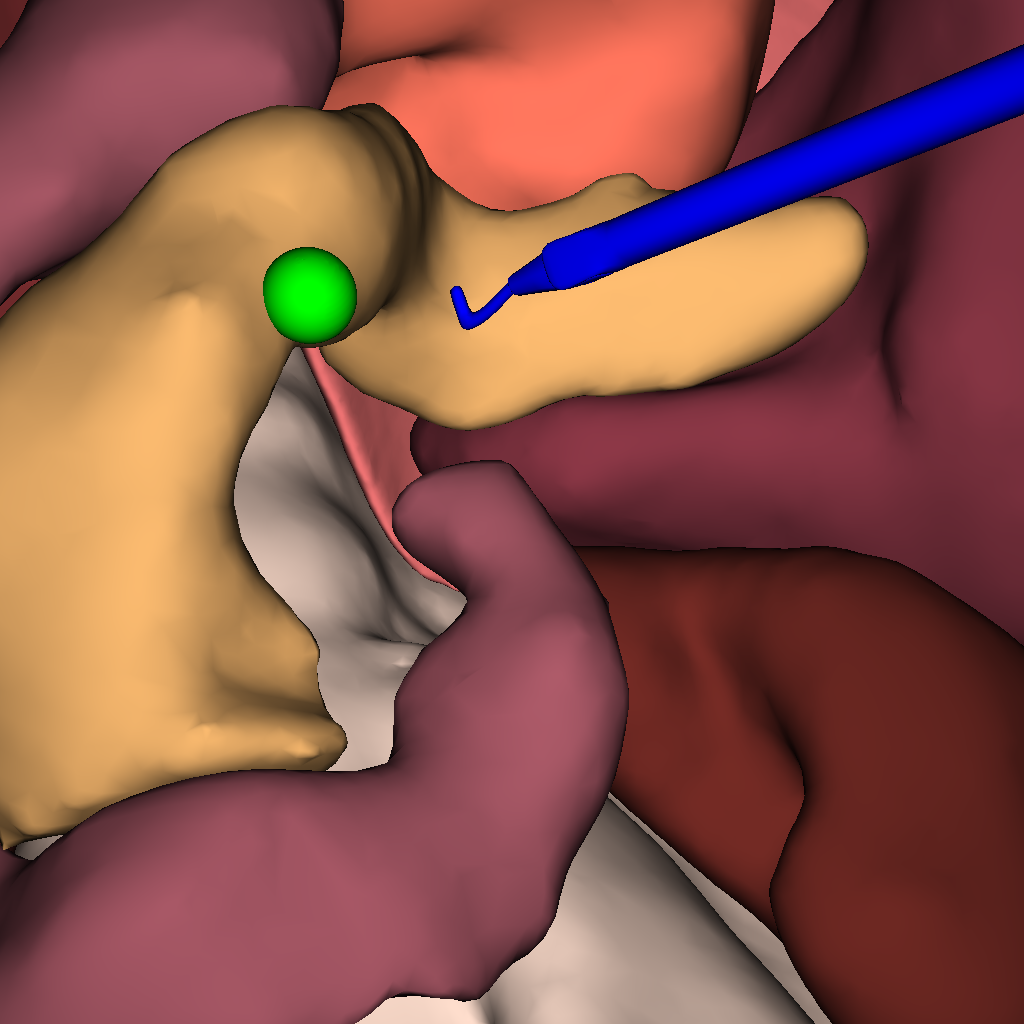}} &
            \hypertarget{env:search}{\textbf{SearchForPointEnv}} contains organ models from the Open Heidelberg Laparoscopy Phantom (OpenHELP) models~\citep{kenngottOpenHelp2015}.
            The instruments in this scene are an oblique-viewing camera with $30^{\circ}$ optics and an electrocautery hook, both controlled in TPSD space.
            The goal is to touch a visual target point on the organ's surface with the tip of the hook to complete the task.
            As the visual observations are limited to what is observed by the camera, the agent has to control the camera to locate both the hook and the target point.
        \end{tabularx}

        ReachEnv is a standard reach task, similar to those in dVRL~\citep{richterDVRL2019} and AMBF-RL~\citep{varierAMBFRL2022}.
        DeflectSpheresEnv introduces the additional challenges of motion in TPSD coordinates around a \gls{rcm}, physical object interaction, and bimanual control of instruments.
        SearchForPointEnv extends the problem of reaching a Cartesian point with a laparoscopic instrument by introducing a controllable camera, thus reframing it as an active vision task.
        Active vision is a core component of laparoscopic surgery, because most tasks are not solvable from a single camera perspective as the region of interest frequently changes over the course of the intervention.
        In most clinical procedures, a surgical assistant is tasked with adjusting the camera position, direction, and zoom level according to the current needs of the operating surgeon.
        Moreover, in laparoscopy, the camera often features an oblique-viewing optic that allows achieving camera perspectives that are unattainable with a forward viewing optic as illustrated in \autoref{fig:oblique}.
        
    \subsection{Deformable Object Manipulation and Grasping Track}
        Deformable object manipulation and grasping involves learning the relation between the actions performed with surgical instruments and their effects on the dynamic behavior of deformable objects.
        In contrast to rigid objects, the effect of an instrument manipulating a deformable object cannot be described by an affine transformation.
        Instead, a complex relationship must be learned between the movements of the agent and the movements of the deformable object.
        The added challenge of grasping can be further subdivided into two tasks: choosing a location for grasping that makes the downstream task solvable, and establishing enough physical contact to enable manipulating the object.
        
        \begin{tabularx}{\textwidth}{lX}
            \raisebox{-\totalheight}{\includegraphics[width=2.5cm]{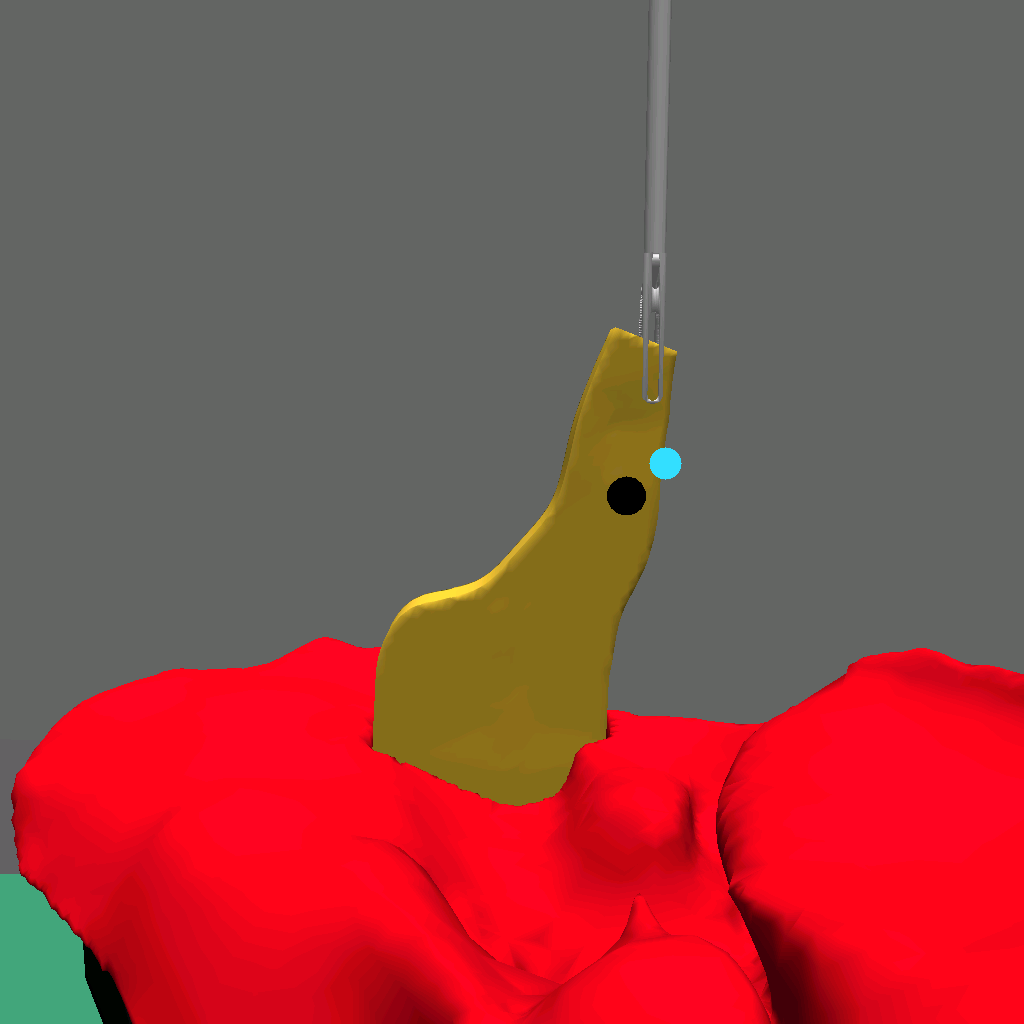}} &
            \hypertarget{env:tim}{\textbf{TissueManipulationEnv}} features yellow, deformable tissue that represents a gallbladder attached to a rigid, red liver.
            The goal is to manipulate the yellow tissue such that a randomly sampled visual landmark on the tissue (black dot) reaches a desired point in the image (blue dot).
            Tissue is manipulated using a laparoscopic grasper, controlled in Cartesian coordinates.
            Initially, the grasper is already attached to a distal point on the tissue, thus omitting the task of grasping.
            The task is finished when the distance between landmark and target in image coordinates is smaller than a given threshold. \\
            \raisebox{-\totalheight}{\includegraphics[width=2.5cm]{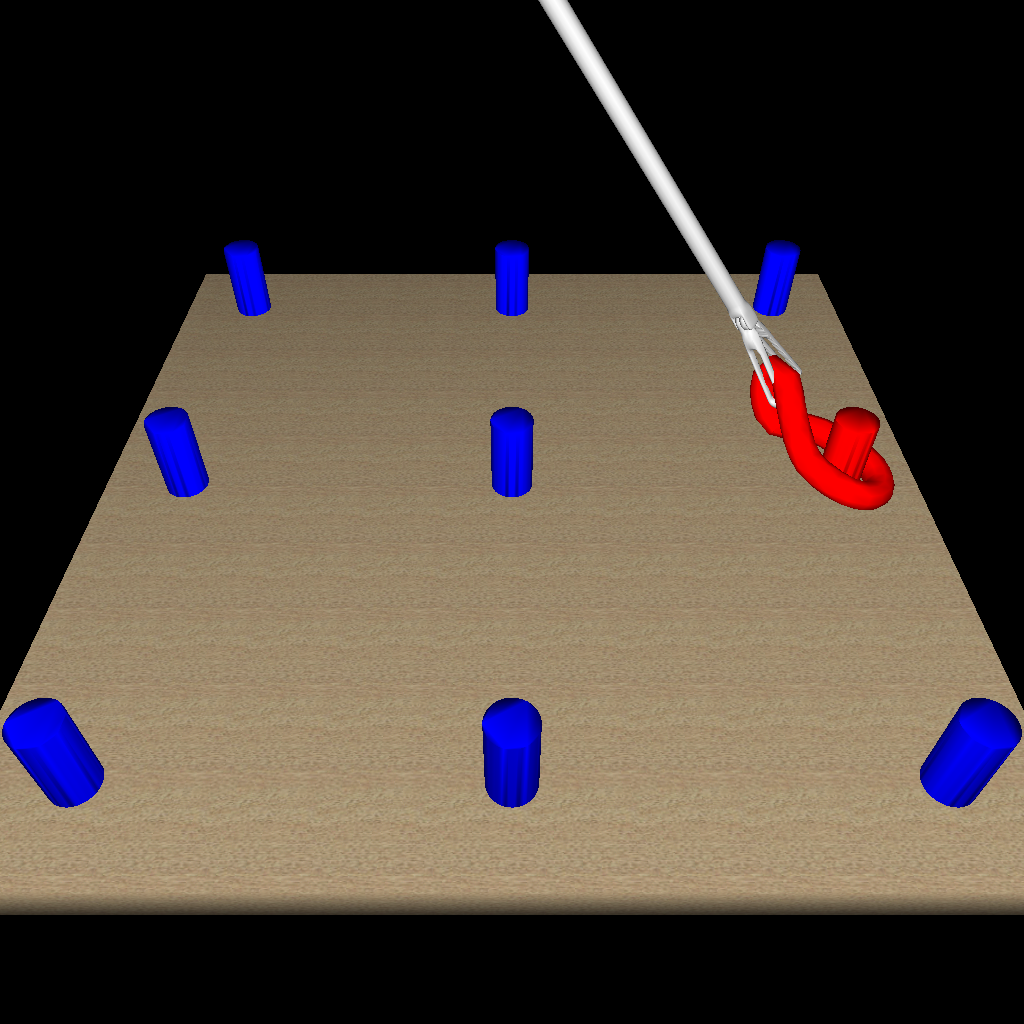}} &
            \hypertarget{env:pick}{\textbf{PickAndPlaceEnv}} is comprised of a board with pegs on a $3\times3$ grid, a deformable torus, and a laparoscopic grasper controlled in TPSD space and jaw angle.
            The goal is to grasp the torus, lift it up to a desired height, and then place it onto the peg with the same color as the torus.
            The task is complete when both the picking and placing phases are finished.\\
            \raisebox{-\totalheight}{\includegraphics[width=2.5cm]{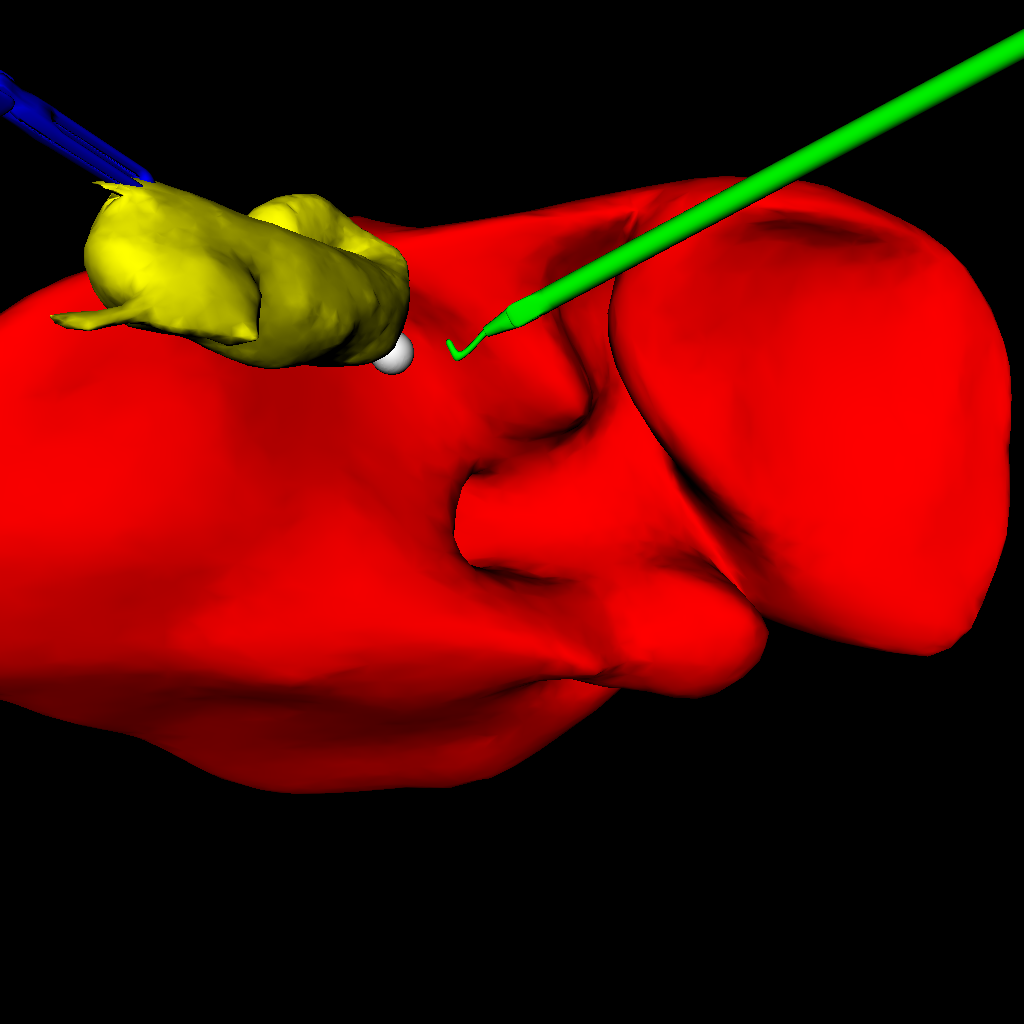}} &
            \hypertarget{env:grasp}{\textbf{GraspLiftAndTouchEnv}} models a sub-task from laparoscopic cholecystectomy, \ie the minimally invasive removal of the gallbladder.
            During dissection of the yellow gallbladder from the red liver, the blue grasper has to grasp the distal end (infundibulum) of the partially resected gallbladder.
            Afterwards, the grasper retracts the gallbladder, exposing a visual marker, which represents the point that should be cut next.
            The green electrocautery hook then navigates to the visual marker and activates in order to cut the tissue.
            The task is complete when the target is visible to the camera and the cauter activates while touching it.
        \end{tabularx}

        TissueManipulationEnv, similar to \citet{shinAunotomousTissue2019}, represents a subtask that is present in various surgical interventions.
        Tissue must be tensioned correctly to make a region of interest visible, make it accessible for cutting, or render the deformations more predictable. 
        PickAndPlaceEnv is a deformable, single-grasper version of the popular peg transfer task of the \gls{fls} program.
        In addition to learning spatial perception, this task requires learning manipulation and control of loose tissue (such as resected lymph nodes) or highly mobile organs (such as the small bowel) in the operating field during laparoscopic surgery).
        GraspLiftAndTouch is an extended version of the LiftAndTouch task from \citet{scheiklCooperativeAssistanceRobotic2021}.
        The task combines the challenges of grasping a deformable object, coordinating heterogeneous instruments, and performing multiple sequential steps to solve the task.
        Although this task models cholecystectomy (\ie, removal of the gallbladder), the task of exposing and then interacting with a region of interest appears often in surgical contexts.
        
    \subsection{Dissection Track}
        Dissection requires learning to remove elements of deformable objects to achieve a desired topological state of the surgical site.
        These topological changes on deformable objects are what distinguish surgery from other medical specializations.
        Unwanted topological changes due to erroneous behavior are generally irreversible and may endanger the patient's health, so precision is required on the part of the agent.
        
        \begin{tabularx}{\textwidth}{lX}
            \raisebox{-\totalheight}{\includegraphics[width=2.5cm]{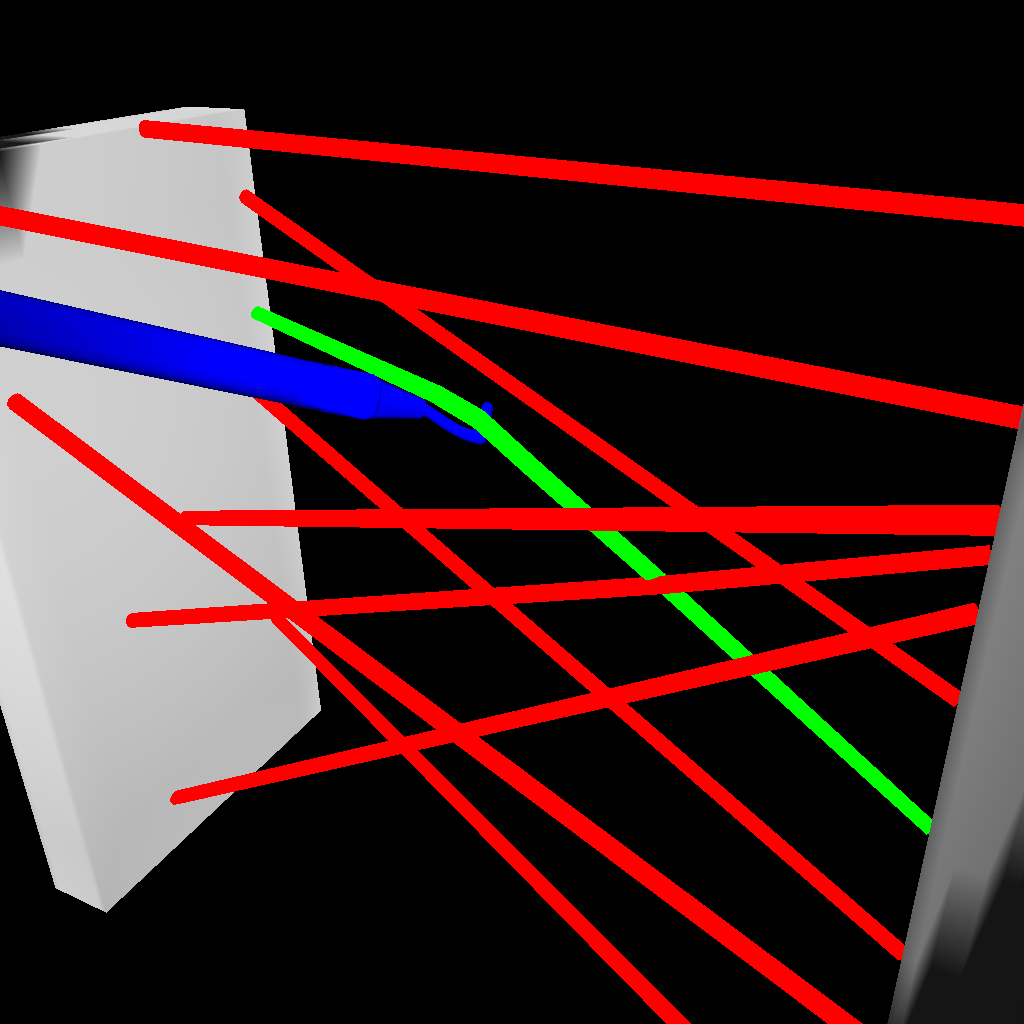}} &
            \hypertarget{env:ropecutting}{\textbf{RopeCuttingEnv}} contains $N$ deformable ropes stretched between two walls.
            An electrocautery hook is controlled in TPSD space to navigate to a highlighted green rope and cut it via activation of the electrocautery hook.
            Subsequently, another rope from the remaining ropes is randomly chosen and marked as active by changing its color to green.
            The task is complete when $M$ active ropes have been cut.
            In contrast to previous environments, the task can fail, if enough incorrect ropes are cut such that cutting $M$ correct ropes is no longer possible.\\
            \raisebox{-\totalheight}{\includegraphics[width=2.5cm]{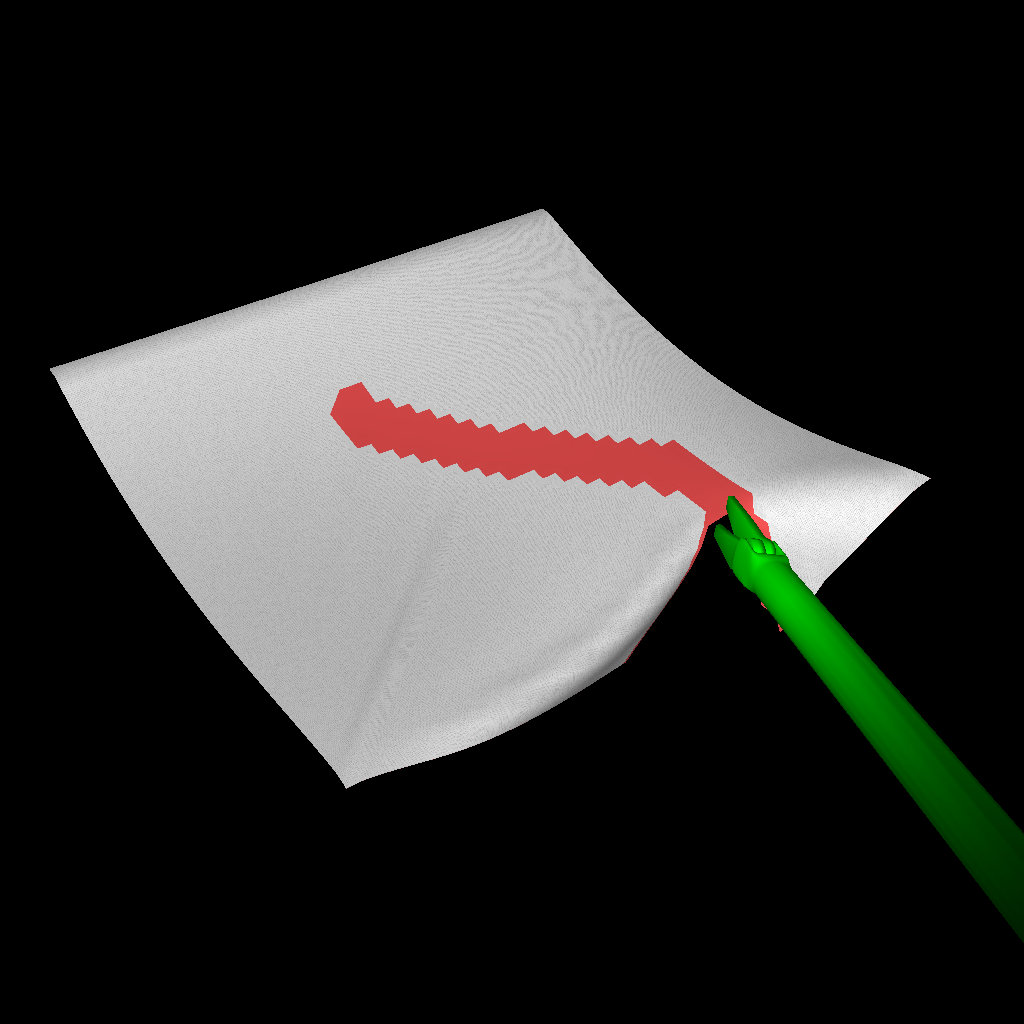}} &
            \hypertarget{env:precision}{\textbf{PrecisionCuttingEnv}} features a deformable cloth on which a desired cutting path is drawn.
            The cloth is fixed at the far edge and the corners near the instrument, to ensure there is enough tension for a cut.
            The agent controls the TPSD state and jaw angle of laparoscopic scissors, and must cut the desired cutting path without damaging the rest of the cloth.
            The path can be parameterized to project linear and sine wave paths onto the cloth.
            The task is complete when $85\%$ of the desired path is cut.\\
            \raisebox{-\totalheight}{\includegraphics[width=2.5cm]{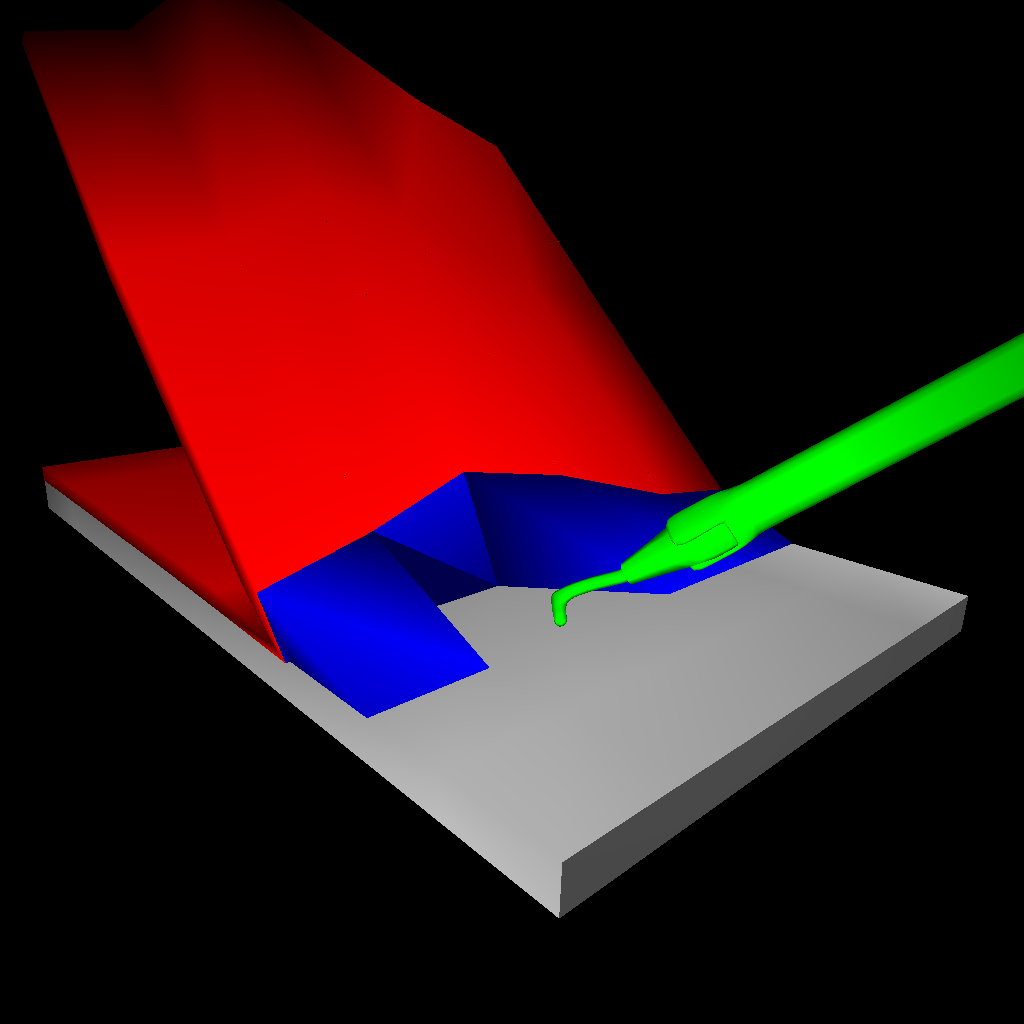}} &
            \hypertarget{env:tissuediss}{\textbf{TissueDissectionEnv}} is comprised of a flap of deformable red tissue that is connected to a rigid board by blue connective tissue.
            The flap is pulled back and away from the board with a constant force, exposing the connective tissue.
            An electrocautery hook is controlled in TPSD space, and is able to cut both connective tissue and flap when it is activated.
            The goal is to cut the connective tissue and dissect the flap from the rigid board without damaging the flap.
            The task is finished when the flap is completely dissected from the board.\\
        \end{tabularx}

        RopeCuttingEnv is based on a common task in surgical training, and is an abstraction of a crucial step in many interventions, for example in laparoscopic cholecystectomy.
        For example, in laparoscopic cholecystectomy, dissecting the gallbladder from the liver requires cutting connective tissue by separating it into filament-like strands and pulling them away from gallbladder and liver.
        These organs should not be cut because (a) damaging the gallbladder leads to leakage of bile into the abdominal cavity which should be avoided to reduce risk of infection, and (b) damaging the liver leads to bleeding, impairs overview of the surgical site, and prolongs the operation.
        PrecisionCuttingEnv is a simplified version of the precision cutting task of the \gls{fls} program, used to teach surgical students how to manipulate tissue to achieve a desired cutting pattern using curved scissors under the constraints of pivotized motion.
        The kind of tissue dissection in TissueDissectionEnv is frequently applied in tissue layer-specific dissection during various procedures and is present in various interventions such as transabdominal preperitoneal inguinal hernia operation (TAPP), laparoscopic cholecystectomy, and total mesorectal excision for rectal cancer (TME).
        
    \subsection{Thread Manipulation Track}
        Thread manipulation involves balancing large motions to reach the overall position of interest with delicate motions to solve the task.
        Threads are difficult to manipulate, due to their high flexibility and length-to-diameter ratio.
        In addition, a thread's shape can only be controlled indirectly, which requires versatile manipulation strategies.
        
        \begin{tabularx}{\textwidth}{lX}
            \raisebox{-\totalheight}{\includegraphics[width=2.5cm]{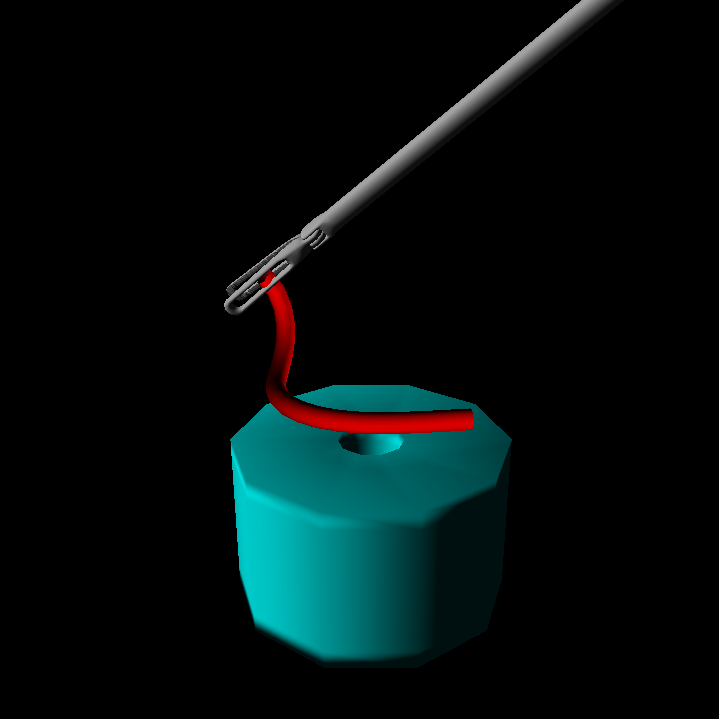}} &
            \hypertarget{env:thread}{\textbf{ThreadInHoleEnv}} consists of a laparoscopic grasper that grasps the upper end of a thread, and a deformable hollow cylinder with a hole.
            The goal is to control the grasper in TPSD space and navigate the hanging end of the thread into the cylinder.
            The task is complete when the thread is inserted to a desired target depth.\\
            \raisebox{-\totalheight}{\includegraphics[width=2.5cm]{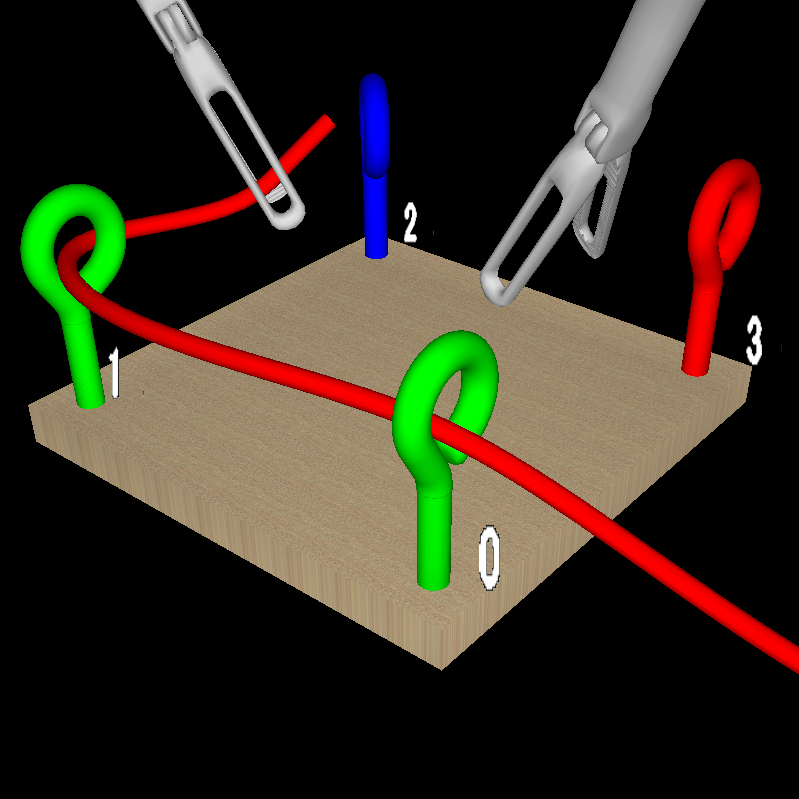}} &
            \hypertarget{env:ropethreading}{\textbf{RopeThreadingEnv}} consists of a board, a set of eyelet screws, a long piece of thread and two laparoscopic graspers.
            The goal is to control both graspers in TPSD space and maneuver the thread through the eyelets in a specific order and direction.
            The task is complete when the rope passes through all eyelets in the correct order.\\
            \raisebox{-\totalheight}{\includegraphics[width=2.5cm]{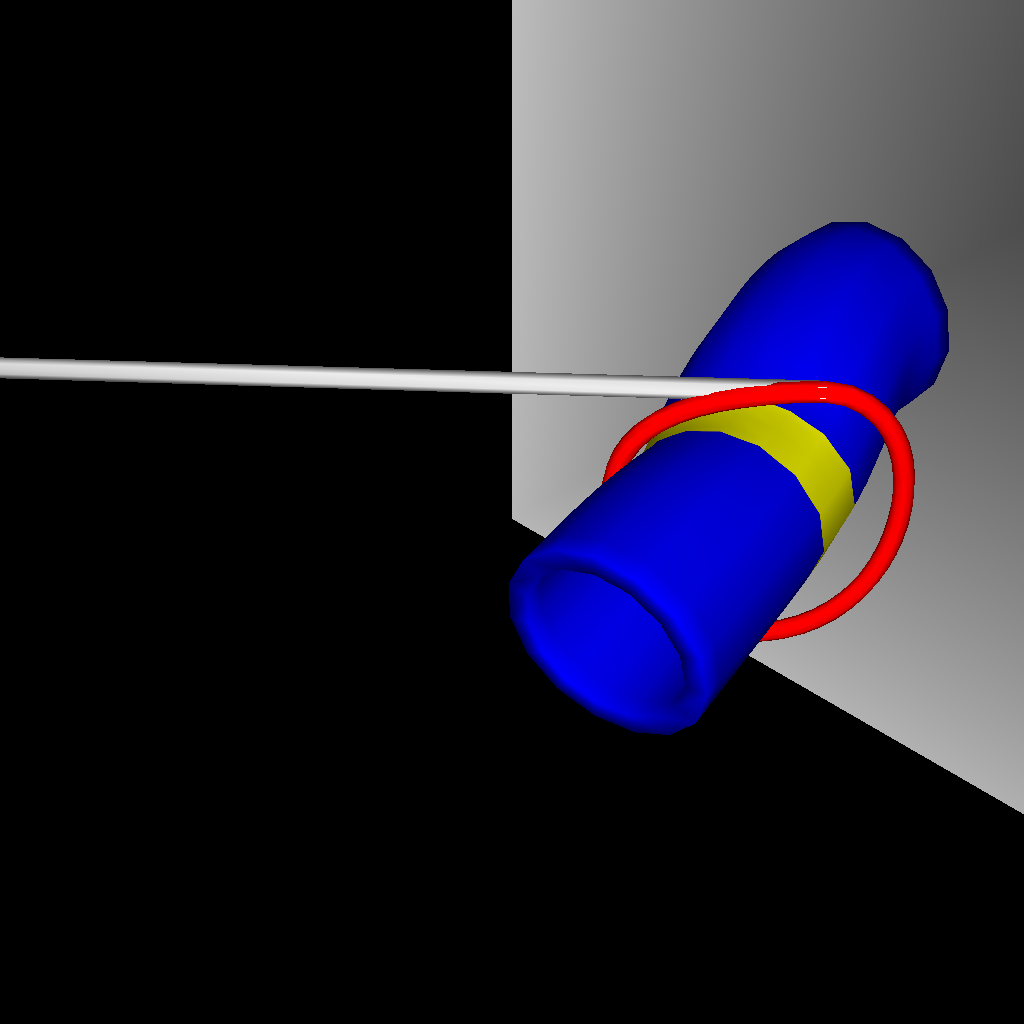}} &
            \hypertarget{env:ligating}{\textbf{LigatingLoopEnv}} consists of a deformable hollow cylinder attached to a wall, and a ligating loop instrument.
            The ligating loop instrument has a deformable loop attached to the end of a rigid shaft.
            The task is to control the instrument in TPSD space and navigate the loop over the cylinder onto the yellow ring, and then close the loop to constrict the cylinder.
            The incremental opening and closing of the loop is controlled through a separate action.
            The task is complete when the loop is closed around the marking to a desired loop length.
        \end{tabularx}

        ThreadInHoleEnv can be freely configured to represent a wide variety of peg-in-hole tasks, from sliding stiff ropes into narrow, deformable cylinders (\eg, urethral catheterization) to indirectly manipulating the distal end of long and flexible ropes into stiff cylinders, simply by changing the dimensions and stiffness of the objects.
        RopeThreadingEnv often used for surgical skill training of novice surgeons, as it builds fine motor control using both hands (\ie, bimanual dexterity) under pivotized motion, and is a prerequisite to suturing.
        Moreover, complex camera guidance may be required if the eyelet screws are facing the camera edge-on, as illustrated in \autoref{fig:oblique} (b).
        LigatingLoopEnv is inspired by the ligating loop task of the \gls{fls} program and requires a high degree of manual dexterity.
        
\section{Reinforcement Learning Experiments}
    \label{sec:experiments}
    \gls{rl} experiments for different configurations of the environments are conducted to establish a learning baseline against which novel methods can be compared.
    Instead of exhaustively tuning the algorithm, hyperparameters, network architecture, and reward functions, the goal of these experiments is to show the effect of various environment configurations on task complexity, while keeping the learning setup constant.
    We train using \glsfirst{ppo}~\citep{schulmanProximalPolicyOptimization2017}, as it is a popular algorithm that has been applied successfully to diverse problems in the literature~\citep{andrychowiczLearningDexterous2020, mirhoseiniGraphPlacement2021, poreLFD2021}, thus making it suitable for use as a baseline.
    All environments were tested in several configurations, using either state or image observations, and varying up to two other environment-specific parameters.
    A complete list of available parameters for environment configuration is given in Appendix B, Table B.
    The experiments span a diverse spectrum of task complexity, from configurations that meet the capabilities of a naive \gls{rl} approach to configurations that exceed them.
    
    This section first presents the individual experiment configurations and their results for each environment grouped by their respective surgical tracks, and then investigates the influence of adding a depth channel or increasing image resolution.
    The frame skip parameter $N$ is set such that the simulated time between observations $\Delta T_o$ is \SI{0.1}{\second}.
    If not stated otherwise, image observations are RGB images with a resolution of $64\times64$.
    A frame stack of $4$ is utilized in both image- and state-based observations.
    Environment-specific hyperparameters, such as time limit for task execution and simulation time step $\Delta T_s$, are reported in Appendix C, Table C.
    Size and content of the hand-crafted state observations are shown in Appendix D, Table D.
    
    \paragraph{Reinforcement Learning Algorithm}
        The StableBaselines3~\citep{raffinStableBaselines3ReliableReinforcement2021} implementation of \gls{ppo} is used with the hyperparameter values reported in \autoref{tab:PPO} across all experiments.
        Each experiment (\ie \gls{ppo} training run) is conducted with $8$ random seeds.
        The training scripts are publicly available in the \sofazoo repository.
        Each training run spans $10^7$ total environment steps, but terminates early if a wall clock time of \SI{48}{\hour} is reached.
        
    \begin{table}[tbh]
        \centering
        \begin{tabular}{l l | l l}
            \toprule
            Hyperparameter & Value & Hyperparameter & Value \\
            \midrule
            total environment steps & $10^7$ & clip range & lin($0.1$)\\
            parallel environments & $8$ & clip range value function & $0.2$\\
            environment steps before update & $8*128$ & value function coefficient & $0.5$\\
            minibatch size &  $256$ & entropy coeffient & $0.0$\\
            update epochs &  4 & maximum gradient norm & $0.5$\\
            discount factor $\gamma$ &  $0.995$ & learning rate & lin($2.5*10^{-4}$)\\
            $\lambda_{GAE}$~\citep{schulmanGAE2015} &  $0.95$\\
            \bottomrule
        \end{tabular}
        \caption{Hyperparameters of \gls{ppo} across all \gls{rl} experiments. lin($x$) denotes a linearly decreasing schedule starting at $x$ and ending at $0.0$ when reaching the total environment steps.}
        \label{tab:PPO}
    \end{table}

    \paragraph{Agent Architectures}
        For state-based observations, the agent consists of two separate neural networks of similar architecture for policy and value estimation.
        Both networks contain two fully connected layers with $256$ neurons each and ReLU non-linearities.
        The policy network has an output layer of $N$ neurons for an action space of $N$ dimensions to predict the mean of a diagonal Gaussian distribution, and a learnable log standard deviation that does not condition on the input.
        The value network has an output layer with a single neuron for value estimation.
        The same network architecture is used in all experiments.
        No parameters are shared and the policy agent is trained from scratch for each experiment.

        For image-based observations, the architecture is adapted to feature convolutional layers for feature extraction.
        Each network processes the image input trough three convolutional layers with square kernel sizes $8$, $4$, $3$ and strides of $4$, $2$, $1$, respectively.
        The output is passed to one fully connected layer with $512$ neurons.
        The heads of policy and value network have the same architecture as in the state-based experiments.
        This is the same architecture as used in \citet{scheiklSimToReal2023}.
        
    \subsection{Spatial Reasoning Track}
        \subsubsection{Configurations}
            \hyperlink{env:reach}{\textbf{ReachEnv}}\hspace{0.5em}
                Parameter $R \in \{\SI{3}{\mm},\ \SI{8}{\mm},\ \SI{20}{\mm}\}$ controls the radius of the target sphere's visual model, and parameter $P \in \{64\times64, 128\times128\}$ controls the resolution of the image observation.
                The state-based task has only one configuration, as the state observation includes both the current end-effector and target positions and thus not depend on the parameters above.
                These configuration parameters are chosen to investigate the influence of the relative visual features sizes.

            \noindent
            \hyperlink{env:deflect}{\textbf{DeflectSpheresEnv}}\hspace{0.5em}
                Parameter $M \in \{1, 2, 5\}$ sets the number of spheres to deflect.
                Parameter $B$ controls whether the agent controls one or two electrocautery hooks.
                The number of spheres on the board is set to $N=5$ across all experiments.
                These configuration parameters are chosen to investigate the influence of the task complexity, and of having to distinguish between both instruments while doubling the size of the action space.
    
            \noindent
            \hyperlink{env:search}{\textbf{SearchForPointEnv}}\hspace{0.5em}
                The environment is tested in 1) the full task of active vision, where the agent controls both camera and electrocautery hook, to touch a highlighted point in the scene, and 2) the reduced task of camera control.
                In the reduced task, the goal is to visualize the highlighted point by centering it in the image observation with a desired distance between camera and point.
                
        \subsubsection{Results}
            \input{figures/learningCurvesSpatialReasoningTrack.tex}
            The learning curves of image- and state-based runs for the environments of the spatial reasoning track are shown in \autoref{fig:results:SpatialReasoningTrack}.
            The state-based policies are able to solve all environments across all configurations. 
            The largest impact on task success is observed for controlling two instead of one instrument in the \hyperlink{env:deflect}{DeflectSpheresEnv}.
            The image-based policies are able to solve the \hyperlink{env:reach}{ReachEnv} for a large visual target sphere of \SI{20}{\mm} radius, and fail for all other configurations and environments.
            The next highest task success is observed for the \hyperlink{env:deflect}{DeflectSpheresEnv} with roughly $30\%$ for the simplest case (one sphere, one instrument), and to roughly $2\%$ for the most complex case (\ie, five spheres, two instruments). 
            The simulation speed of \hyperlink{env:search}{SearchForPointEnv} is lower than the other environments, because collision detection and rendering is costly for scenes with multiple complex shapes.
            Thus, the image based runs were ended by the time criterion of \SI{48}{\hour} per run and did not reach the desired $10^7$ environment steps.
            Both variants of \hyperlink{env:search}{SearchForPointEnv} inherently rely on image observations.
            Thus, the state-based experiments should be regarded as supplementary information to validate that the reported performance for the image-based experiments is not due to reward factor misspecification.

    \subsection{Deformable Object Manipulation and Grasping Track}
        \subsubsection{Configurations}
            \hyperlink{env:tim}{\textbf{TissueManipulationEnv}}\hspace{0.5em}
                Threshold $T \in \{\SI{2}{\mm},\ \SI{5}{mm}\}$ controls the distance between landmark and target point in image space required to complete an episode.
                Parameter $N$ sets whether the landmark is a fixed point on the tissue, or sampled from a set of points on the tissue after each environment reset.
                These configuration parameters are chosen to investigate whether the agent is able to learn accurate control of the dynamic behavior of the deformable tissue for multiple points on the tissue.

            \noindent
            \hyperlink{env:pick}{\textbf{PickAndPlaceEnv}}\hspace{0.5em}
                The environment is tested with 3 different combinations of phases: \textit{Pick}, \textit{Place}, \textit{Pick and Place}.
                For configuration \textit{Pick}, the episode ends when the torus is grasped and the instrument reaches the desired lifting height.
                For configuration \textit{Place}, the episode starts with torus already grasped and instrument at desired lifting height, and ends when the torus is placed on the active peg.
                Additionally, parameter $M$ controls the material parameters of the torus to model either a stiff or a soft torus.
                For the stiff case, the mechanical beam radius of the torus' \gls{fem} model is increased by a factor of $5$ while the mass is decreased by a factor of $2.5$.
                The phase combinations investigate the relative complexity of the phases and the added difficulty of a multi-phase task.
                The material parameters are investigated to show the difference between manipulating highly deformable and more stiff objects.

            \noindent
            \hyperlink{env:grasp}{\textbf{GraspLiftAndTouchEnv}}\hspace{0.5em}
                The environment is tested in six configurations with different combinations of phases: \textit{Gr}, \textit{GrLi}, \textit{GrLiTo}, \textit{Li}, \textit{LiTo}, \textit{To}.
                Phase names Grasp, Lift, and Touch are abbreviated as \textit{Gr}, \textit{Li}, and \textit{To}, respectively.
                For configuration \textit{GrLi}, the task is finished grasping and lifting phases are complete.
                For configuration \textit{LiTo}, the task starts with the gallbladder already grasped, and ends when the electrocautery hook correctly activates in the target point.
            
        \subsubsection{Results}
            \input{figures/learningCurvesDeformableObjectManipulationTrack.tex}
            The learning curves of image- and state-based runs for the environments of the  deformable object manipulation and grasping track are shown in \autoref{fig:results:DeformableObjectManipulationTrack}.
            The state-based policies are able to solve all environments in most configurations.
            The mechanical parameters of \hyperlink{env:pick}{PickAndPlaceEnv} influence the task success for pick and place phases differently.
            The highest success rate of almost $100\%$ is reached for picking up a soft torus.
            Picking up the stiff torus reaches only $90\%$ task success, same as placing the stiff torus.
            Placing the soft torus reaches a maximum of $80\%$ task success at $3$ million steps and then declines to $60\%$ before recovering.
            Task success decreases with the number of phases that have to be learned by the policy.
            Learning all phases reaches a success rate of roughly $40\%$ and $8\%$ for the \hyperlink{env:grasp}{GraspLiftAndTouchEnv} and \hyperlink{env:pick}{PickAndPlaceEnv}, respectively.
            Without randomizing the position of the point on the tissue in \hyperlink{env:tim}{TissueManipulationEnv}, the image-based policy is also able to learn the task with roughly $100\%$ task success.
            Lowering the threshold for task completion and randomizing the position both slow down learning.
            The \SI{5}{mm} threshold configuration reaches a success rate of roughly $95\%$, while the \SI{2}{mm} threshold configuration converges to $40\%$.
            For \hyperlink{env:pick}{PickAndPlaceEnv}, the image-based policy, in contrast to the state-based policy, picking up the stiff torus has a higher success rate than picking up the soft torus.
            The image-based policy is unable to learn the full task with all phases of the \hyperlink{env:grasp}{GraspLiftAndTouchEnv}.
        
    \subsection{Dissection Track}
        \subsubsection{Configurations}
            \hyperlink{env:ropecutting}{\textbf{RopeCuttingEnv}}\hspace{0.5em}
                Parameter $R \in \{5, 10\}$ controls how many random ropes are generated between the walls at each environment reset.
                Parameter $C \in \{1, 3\}$ sets the number of active ropes that must be cut to complete the episode.
                With more ropes in the scene, navigating to and cutting the marked rope becomes increasingly difficult.
                However, this also means that more ropes can be cut incorrectly before the task is failed.
                For example, with $R=5$ and $C=3$, it is not possible to complete the task if more than two incorrect ropes are cut.

            \noindent 
            \hyperlink{env:precision}{\textbf{PrecisionCuttingEnv}}\hspace{0.5em}
                The function projected onto the cloth to mark the desired cutting path is either a linear function $l(x) = ax + b$ or a sine function $s(x) = c\sin{fx} + d$, where $x$ is along the depth dimension of the cloth.
                The parameters for these functions are either fixed or sampled randomly for each trajectory.
                The specific function controls how complex the desired cutting path is, while randomizing parameters of the path forces the agent to condition on the observed desired cutting path, instead of solving the task by \textit{memorizing} a specific trajectory.
                For the fixed case, the parameters are $a=0$, $b=0.5$, $c=15$, $d=0.6$, and $f=\frac{1}{75}$, and $x$ is limited to the interval $[0, 0.5]$.
                For the random case, the parameters are uniformly sampled as follows: $a \sim U(-0.5, 0.5)$, $b \sim U(0.3, 0.7)$, $c \sim U(10, 20)$, $d \sim U(0.3, 0.7)$, $f \sim U(\frac{0.5}{75}, \frac{1.5}{75})$, and $x$ is limited to the interval $[0, x_{high}]$, where $x_{high} \sim U(0.5, 0.7)$.
                The values are unitless and represent coordinates relative to the size of the cloth.

            \noindent 
            \hyperlink{env:tissuediss}{\textbf{TissueDissectionEnv}}\hspace{0.5em}
                Parameter $R \in \{2, 4\}$ represents the number of rows of connective tissue must be cut to complete the task.
                Parameter $V$ controls whether a visual indicator marking the point on the connective tissue that is closest to the electrocautery hook's tip is enabled or disabled.
                While $R$ controls the overall complexity of the task, the visual marker represents an initial guide for task-driven visual feature extraction.
            
        \subsubsection{Results}
            \input{figures/learningCurvesDissectionTrack.tex}
            The learning curves of image- and state-based runs for the environments of the dissection track are shown in \autoref{fig:results:DissectionTrack}.
            The difference in the number of total environment steps between policies that learn quickly and those that do not is caused by the time required to reset environments that feature topological changes.
            Such a reset requires creating a process at the operating system level and initializing a new simulation instance.
            Better policies cause environment resets more often, resulting in an overall decrease in sampled steps per second by a factor of $3$ to $10$.
            For example in \hyperlink{env:ropecutting}{RopeCuttingEnv}, a successful policy is able to complete the easiest configuration in $40$ steps, which results in $10$ times more resets compared to an unsuccessful policy that times out after $400$ steps.
            Consequently, learning runs of successful policies are more often terminated by the time limit of \SI{48}{\hour} than by the limit of total environment steps.
            
            For \hyperlink{env:ropecutting}{RopeCuttingEnv} the number of total ropes in the scene has a higher impact on task success than the number of ropes that should be cut.
            Compared to the state-based policies, the image-based policies reach much lower task success rate with roughly $40\%$ for the simplest (cut $1$ rope out of $5$) and $0\%$ for the most complex (cut $3$ ropes out of $10$) configuration.
            For \hyperlink{env:precision}{PrecisionCuttingEnv}, introducing noise to the function parameters strongly impacts the final task success rate and learning speed.
            The final success rate drops from around $95\%$ for the non-randomized case to $50\%$ for the image- and $60\%$ for the state-based policies in the randomized case.
            State-based policies are not able to solve the \hyperlink{env:tissuediss}{TissueDissectionEnv} and reach a final task success rate of less than $10\%$.
            Image-based policies learn the task to a mean success rate of roughly $60\%$, albeit with a large variance across runs.
            \hyperlink{env:precision}{PrecisionCuttingEnv} and \hyperlink{env:tissuediss}{TissueDissectionEnv} are the first environments where image-based policies were able to compete with, and even surpass the success rate of state-based policies.
    
    \subsection{Thread Manipulation Track}
        \subsubsection{Configurations}
\begin{figure}[tbh]
                    \centering
                    \begin{subfigure}[b]{0.15\textwidth}
                        \centering
                        \includegraphics[width=\textwidth]{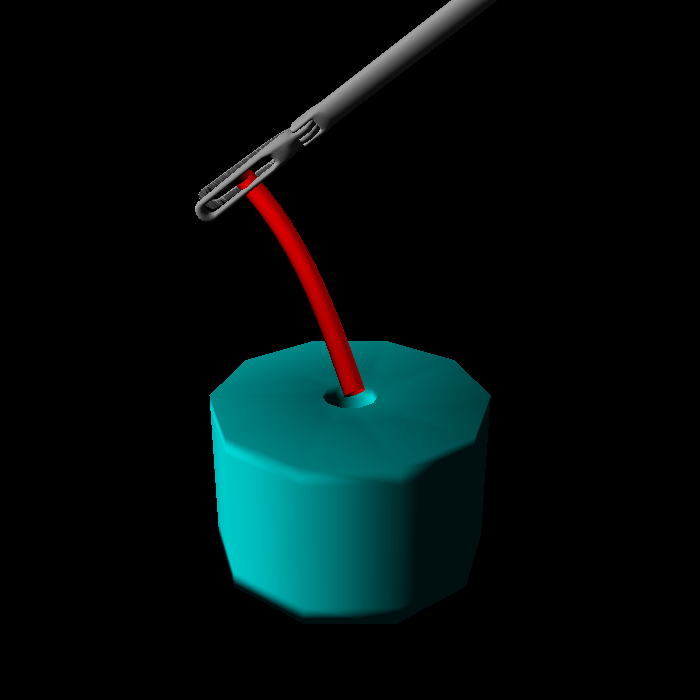}
                        \caption{}
                    \end{subfigure}
                    \begin{subfigure}[b]{0.15\textwidth}
                        \centering
                        \includegraphics[width=\textwidth]{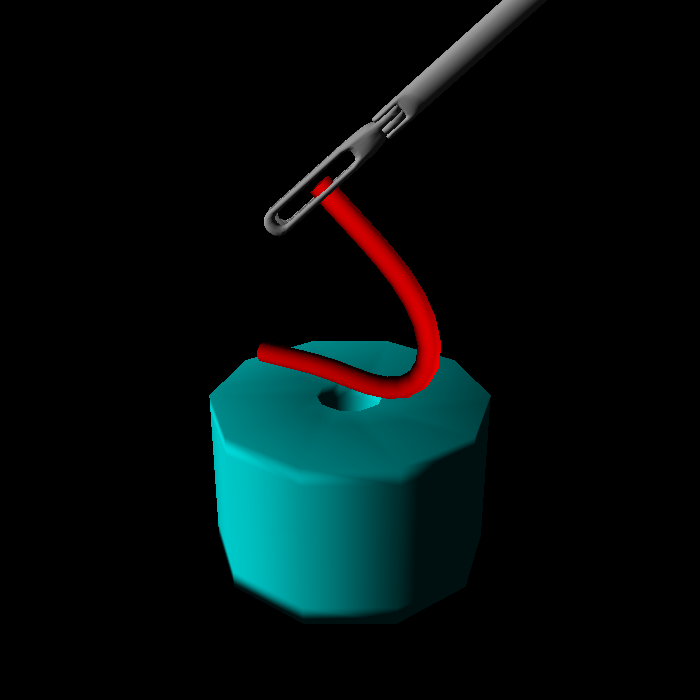}
                        \caption{}
                    \end{subfigure}
                    \begin{subfigure}[b]{0.15\textwidth}
                        \centering
                        \includegraphics[width=\textwidth]{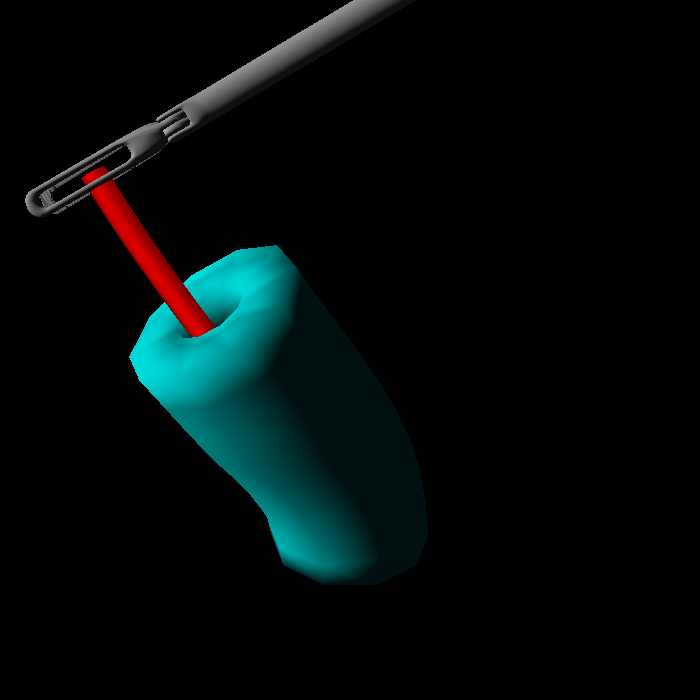}
                        \caption{}
                    \end{subfigure}
                    \hfill
                    \begin{subfigure}[b]{0.45\textwidth}
                        \centering
                        \includegraphics[width=0.3\textwidth]{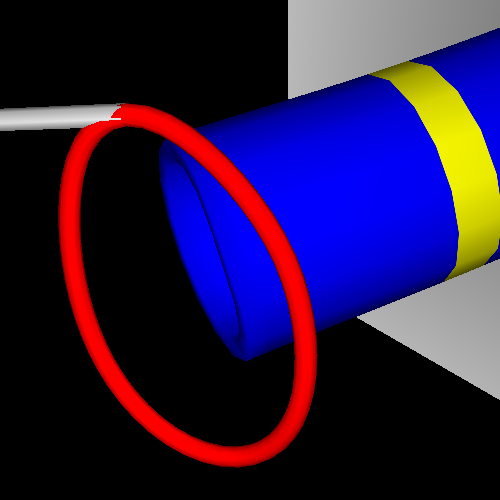}
                        \includegraphics[width=0.3\textwidth]{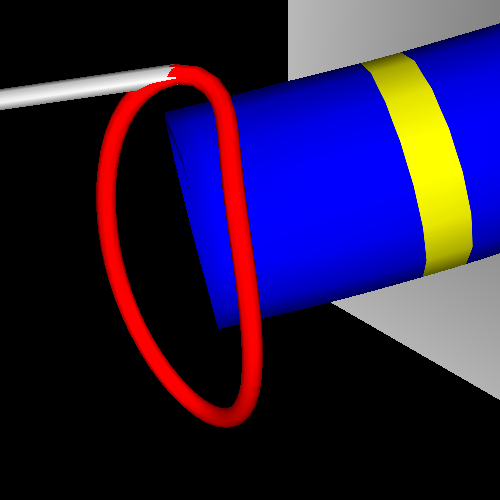}
                        \includegraphics[width=0.3\textwidth]{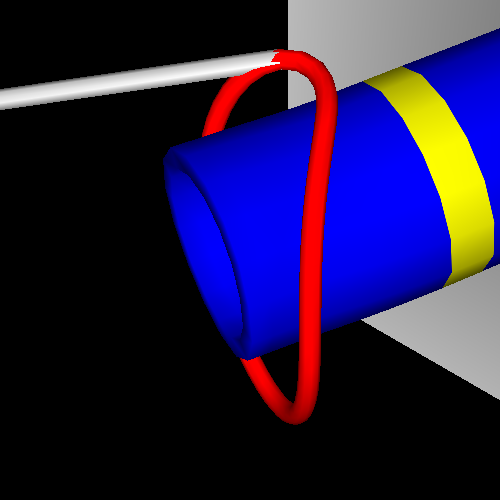}
                        \caption{}
                    \end{subfigure}
                    \caption{
                        ThreadInHoleEnv configurations of the mechanical parameters for the a) \textit{normal}, b) \textit{flexible}, c) \textit{inverted} case.
                        d) possible strategy of navigating the loop over the cavity in the \textit{soft} case of the LigatingLoopEnv.
                        The interactions with cavity are necessary to implicitly control the shape of the loop and finish the task.
                    }
                    \label{fig:ThreadInHoleConfigs}
                \end{figure}

            \hyperlink{env:thread}{\textbf{ThreadInHoleEnv}}\hspace{0.5em}
                Parameter $M \in \{\text{\textit{normal}},\ \text{\textit{flexible}},\ \text{\textit{inverted}}\}$ controls the mechanical properties of thread and hole as illustrated in \autoref{fig:ThreadInHoleConfigs} (a)-(c).
                The \textit{flexible} case has a longer and more flexible thread, compared to the \textit{normal} case.
                The agent thus has to learn a more complex dynamical behavior of the thread.
                The \textit{inverted} case is challenging because the hole must be deformed with the indirectly controlled tip of the thread to insert, due to the constrained motion of the pivotized grasper.
                Parameter $N$ controls whether the camera pose is randomized for each trajectory.
                This investigates whether the agent is able to learn a visual feature extraction that is robust to changes in camera perspective.

            \noindent
            \hyperlink{env:ropethreading}{\textbf{RopeThreadingEnv}}\hspace{0.5em}
                For all configurations, the number of loops on the board is reduced to a single one, as preliminary experiments with more than one loop were unsuccessful.
                Parameter $M \in \{\text{\textit{base}},\ \text{\textit{bimanual}}\}$ controls the criterion that marks the episode as done.
                In the \textit{base} case, the episode is marked as done, when $5$\% of the rope is passed through the loop.
                In the \textit{bimanual} case, the episode is marked as completed, when the rope is passed through the loop, and grasped by the second grasper on the other side of the loop.
                The \textit{bimanual} case adds a phase to the task that brings it closer to behavior that can solve the iterative threading and regrasping required to solve the task with more loops on the board.
                Parameter $N$ controls whether noise is added to the position and orientation of the loop at each environment reset.
                Each episode starts with the rope grasped by the right grasper between $10\%$ and $30\%$ of the rope length from the tip of the \SI{200}{mm} long rope.
                So even in the case where the loop pose is not randomized, a \textit{memorized} trajectory is not able to solve the task.

            \noindent
            \hyperlink{env:ligating}{\textbf{LigatingLoop}}\hspace{0.5em}
                The environment is tested in configurations $C \in \{\text{\textit{soft}},\ \text{\textit{stiff}}\}$ that control the mechanical parameters of the loop.
                The mechanical behavior of the loop heavily influences the difficulty of the task.
                A stiff loop maintains its overall round shape, so navigating the loop over the cavity is easier compared to the soft case.
                In the soft case, the agent has to learn a strategy that uses the cavity to control the shape of the loop through physical interaction, in order to make navigating the loop over the cavity possible as illustrated in \autoref{fig:ThreadInHoleConfigs} (d).
        
        \subsubsection{Results}
            \input{figures/learningCurvesThreadManipulationTrack.tex}
            The learning curves of image- and state-based runs for the environments of the thread manipulation track are shown in \autoref{fig:results:ThreadManipulationTrack}.
            State-based policies are able to learn all tasks across all configurations, except for \hyperlink{env:ligating}{LigatingLoopEnv} ($20\%$ and $0\%$ task success) and the bimanual configuration of \hyperlink{env:ropethreading}{RopeThreadingEnv} ($0\%$ task success).
            The image-based policies deliver comparable results, even outperforming the state-based policy on \hyperlink{env:ligating}{LigatingLoopEnv} in the stiff configuration with $40\%$ task success.
            The image-based policies are unable to solve the task for randomizing camera pose in \hyperlink{env:thread}{ThreadInHoleEnv} or loop pose in \hyperlink{env:ropethreading}{RopeThreadingEnv}.
            The simulation speed of \hyperlink{env:ligating}{LigatingLoopEnv} is much slower compared to the previous environments, limiting the total number of environment steps to under $10^6$.
            Constricting the cavity with a loop is very computationally expensive for both collision detection and deformation including complex constraint resolution.

    \subsection{Image Resolution}
        \label{subsec:stateVSimage}
        We additionally test the effect of image resolution on environments where precision might be limited by low-resolution images.
        In \hyperlink{env:reach}{ReachEnv}, the configurations with a higher resolution of $128\times128$ do not outperform the lower resolution configurations.
        When increasing the resolution in \hyperlink{env:deflect}{DeflectSpheresEnv}, the success rate increases by roughly $5\%$, however, only in the configurations with one instrument.
        The most notable difference occurs for \hyperlink{env:tim}{TissueManipulationEnv}, where a $128\times128$ resolution with a threshold of \SI{2}{\mm} reaches a task success rate of $100\%$, compared to only $40\%$ for a resolution of $64\times64$.
        For a threshold of \SI{5}{mm}, the final success rate is the same but learning is faster with a resolution of $128\times128$.
    
    \subsection{Depth Information}
        We additionally investigate the effect of adding depth information to image observations in all environments, as this greatly increases the information available to the agent.
        We use the same configurations as for RGB image-based policies, yet adding a depth channel to create RGBD images.
        Across environments, adding depth information usually resulted in only minor increases in success rate of around $5\%$ to $10\%$.
        These results are aligned with the findings of~\cite{barnoyLeanRL2021} that compare various different image-based observation types on a reach and a suturing task in \gls{rals}.
        The learning curves for these experiments on the spatial reasoning track are shown in \autoref{fig:results:DepthInformation}.
        \hyperlink{env:reach}{ReachEnv} shows the most noticeable difference between RGB and RGBD observations.
        For a radius of \SI{20}{mm}, the policy learns faster and reaches a higher final success rate, and this improvement is also observed for a radius of \SI{8}{mm}, but only for the higher resolution variant.
        \input{figures/learningCurvesRGBD.tex}

\section{Discussion}
    \label{sec:discussion}
    In this section, we show how our results reflect several well-known open problems in \gls{rl} research.
    We present overall trends in the environment parameters with the largest effect on task complexity, which include using high-dimensional image observations instead of compact state observations, cooperation between multiple agents, and multi-phase tasks.
    We also discuss reward design and learning policies with safety constraints, two prominent topics in the community that are particularly relevant to surgical robotics.
    By showing that these open problems are represented throughout the environment suite, we substantiate that LapGym is an appropriate proving ground for further research into these problems.
    
    \paragraph{State vs Image Observations}
        Across tested environments and configurations, state-based policies usually outperform their image-based counterparts.
        However, the opposite is observed for \hyperlink{env:tissuediss}{TissueDissectionEnv} and \hyperlink{env:ligating}{LigatingLoopEnv}, where image-based policies reach $30\%$ and $60\%$ greater task success, respectively.
        Additionally, state observations did not perform much better than image observations on \hyperlink{env:precision}{PrecisionCuttingEnv}.
        These environments require localizing a pattern or a piece of tissue that is deformed in a complex way, making it difficult to capture the relevant information with a hand-crafted state vector based on a limited number of points.
        In these cases, state-based policies must deal with uncertainty about their environment, where image-based policies do not.
        As environments and tasks increase in complexity through greater deformability and more topological changes, the capabilities of state-based approaches may be limited, motivating the shift towards image-based policies in robotic surgery~\cite{scheiklSimToReal2023}.

        However, image observations can be non-Markovian on first-person navigation tasks.
        For example, in \hyperlink{env:search}{SearchForPointEnv}, the agent can get lost if it reaches a position where it can only see the abdominal wall or some other featureless surface, as it can only condition on the most recent observation.
        Similarly, in the multi-instrument variant, the electrocautery hook is difficult to locate if it moves off frame, especially because it does not necessarily stay at its last seen position.
        Short of using a recurrent policy architecture, proprioception or sensor fusion can also be utilized to mitigate this failure mode.

    \paragraph{Image Resolution}
        The optimal resolution for a particular task depends on the sizes of relevant objects in the scene and the sizes of the kernels in the agent's convolutional layers.
        In \hyperlink{env:tim}{TissueManipulationEnv}, the relevant features are the landmark and target points, as well as the distance between them.
        Because the sizes of these points are well matched to the kernel size, increasing the resolution provides the agent with more information and increases success rate.
        Note that this is only the case for randomized landmark points, since otherwise the policy can simply learn to manipulate the tissue as a whole without paying attention to the specific landmark.

        In \hyperlink{env:reach}{ReachEnv}, not only is the target sphere relevant, but also the relative depths of the target sphere and the end-effector.
        Estimating this depth is only possible through minute changes in the sizes of the spheres.
        Because the kernel sizes appear to be mismatched for features of this scale, increasing the resolution does not significantly change the outcome, other than slowing down learning due to a larger model size.
        Optimizing the network parameters may improve success rate on this task, but is outside of the scope of this work.
        
    \paragraph{Multi-Instrument Collaboration}
        The presented environments require different types of collaboration between instruments to solve the tasks at hand.
        Sequential coordination, where only one instrument is \textit{active} at a time, is present in \hyperlink{env:grasp}{GraspLiftAndTouchEnv}, \hyperlink{env:ropethreading}{RopeThreadingEnv}, and \hyperlink{env:deflect}{DeflectSpheresEnv}.
        Simultaneous coordination, where multiple instruments are \textit{active} at the same time, is present in the active vision task of \hyperlink{env:search}{SearchForPointEnv}, the two-instrument variants of \hyperlink{env:precision}{PrecisionCuttingEnv} and \hyperlink{env:ligating}{LigatingLoopEnv}, as well as the variant of \hyperlink{env:tissuediss}{TissueDissectionEnv} where the force pulling back on the tissue is controllable.
        Initial investigations with these environments showed that simultaneous coordination is significantly harder to learn than sequential coordination.
        
        Simultaneous coordination can exacerbate the problem of credit assignment.
        For example, in the active vision task of \hyperlink{env:search}{SearchForPointEnv}, actions that control the camera are uncorrelated with the shaped reward, which only depends on the position of the movement of the electrocautery hook.
        Although it is possible to add further shaped rewards that depend on the camera movement, reward engineering is at best tedious and at worst can lead to a bias in task execution or suboptimal learning.
        We will add more tasks that require simultaneous coordination to \sofaenv, in order to better support research into algorithms for this particular challenge, for example multi-agent \gls{rl}.

    \paragraph{Multi-Phase Tasks}
        Many surgical procedures are comprised of several distinct phases, often at multiple levels of granularity.
        In \sofaenv, multi-phase tasks include the single-instrument task of \hyperlink{env:pick}{PickAndPlaceEnv} and the multi-instrument task of \hyperlink{env:grasp}{GraspLiftAndTouchEnv}.
        When learned by a single agent, multi-phase tasks are comparable to learning a task with sequential coordination.
        Designing reward functions for such tasks is challenging, and learning is increasingly difficult as more phases are added.
        Alternative approaches to the general problem of multi-phase tasks include multi-agent \gls{rl}~\citep{scheiklCooperativeAssistanceRobotic2021}, hierarchical \gls{rl}~\citep{pateriaHierarchical2021}, or meta-learning~\citep{yuMetaWorld2020}.
        Multi-agent and hierarchical \gls{rl} may address the challenge by learning specific policies and their coordination for each phase, while meta-learning may view a multi-phase task as different tasks from the same family.
        
    \paragraph{Reward Hacking}
        \gls{rl} algorithms generally learn faster when given shaped rewards~\citep{pathakCuriosityDriven2017}.
        Recent algorithms are able to learn from sparser rewards~\citep{hafnerDeepHierarchical2022} but are often more computationally expensive.
        A practical challenge with shaped rewards is \textit{reward hacking}, where the policy is able to maximize the return without actually solving the task.
        For example, \hyperlink{env:ligating}{LigatingLoopEnv} contains a reward factor for the overlap between the loop and the target marking, intended to help the policy learn this subtle difference between successful and unsuccessful trajectories.
        If this reward factor is set too high, the policy  only tries to maximize the overlap instead of solving the task, as this would end the episode early and decrease the total return.
        Time step costs and control costs may mitigate this issue in this case, but also introduce more reward factors that must be tuned correctly.
        Curriculum learning~\citep{narvekarCurriculumLearning2022} addresses this problem by incrementally reducing the magnitude of the shaped rewards throughout training, ultimately leaving only the sparse rewards that truly represent task success.
        
    \paragraph{Safety Constraints} 
        Especially in a surgical context, it is desirable for an \gls{rl} agent to learn to avoid certain unsafe behaviors as it completes the task~\citep{poreSafeRL2021}.
        Although these undesired behaviors cause no real damage while learning in simulation, they are unacceptable during task execution.
        Punishments (\ie, negative rewards) are the typical solution for this scenario, however, care must be taken to ensure that exploration is not hindered.
        This balance is especially precarious if desired and undesired states appear very similar in observation space, for example in \hyperlink{env:precision}{PrecisionCuttingEnv}.
        One possible solution to this challenge is curriculum learning~\cite{scheiklSimToReal2023}, where punishments for safety violations are incrementally increased throughout training, without curtailing initial exploration.
        
        Early termination of the trajectory upon violation of safety constraints made learning more likely to fail in our early experiments, presumably because it makes reward hacking easier.
        If rewards are negative in expectation, it can be advantageous to accept a one-time punishment that ends the episode instead of allowing negative rewards to accumulate.
        Instead, in all environments with the exception of \hyperlink{env:ropecutting}{RopeCuttingEnv}, only successful task completion ends an episode early, and the discount factor incentivizes the agent to solve the task faster.

    \subsection{Technical Limitations and Future Roadmap}
        \paragraph{LapGym as a Benchmark}
            In the near future, we intend to host a challenge where research groups can submit their trained \gls{rl} models to be evaluated by the challenge host.
            Prior to this, we will make the code open source and allow the community to decide on a relevant set of standard configurations for the challenge.
            A set of human expert trajectories will also be provided to serve as a baseline and as data for imitation learning approaches.
            Furthermore, new features will be added incrementally as the need arises in the community and in our own research.
            As Gym is no longer maintained, we will port LapGym to its successor, Gymnasium~\citep{faramaGymnasium2022}, as soon as compatibility for it is added to Stable Baselines~3.
            
        \paragraph{Simulation Speed}
            Environments that simulate complex deformations of large volumes are highly computationally demanding.
            This may especially be a limitation for on-policy \gls{rl} algorithms, which typically have lower sample efficiency and require more environment steps.
            SofaCuda is an active effort to implement GPU-based physics simulation as a drop-in replacement for CPU-based \gls{sofa} components.
            
        \paragraph{Realistic Rendering}
            \gls{sofa} focuses on fast and accurate \gls{fem} simulation.
            The rendering capabilities of \gls{sofa} are thus rather limited and cannot compete with modern rendering pipelines, like in the Unity game engine, for example.
            For generating point cloud and semantically segmented image observations, \sofaenv already uses the rendering tools of Open3D.
            The same approach may be employed to outsource rendering of RGB images to an external library, such as Pytorch3D.
        
        \paragraph{Haptic Feedback for Human Control}
            \gls{sofa} is used for training surgeons in virtual reality settings with detailed force feedback~\citep{courtecuisseHapticRendering2015}.
            We plan to include haptic feedback in the environments as additional sensor feedback for the learning algorithms as well as to extend support for teleoperation.
            The aim of this extension is to contribute to the democratization of surgical skill via the tactile internet~\citep{fitzekTactileInternet2021}.

\section{Conclusion}
    \label{sec:conclusion}
    This work proposes LapGym, an open-source environment framework for \gls{rl} in \gls{rals}.
    The framework contains $12$ challenging and highly parametrizable \gls{rl} environments that are adapted to the unique requirements of \gls{rals}.
    Baseline \gls{rl} experiments with \gls{ppo} across different configurations of the environments show the limitations of current \gls{rl} methods for learning clinically relevant skills in \gls{rals}.
    The framework provides researchers with utilities to rapidly create new \gls{rl} environments, collect expert data for imitation learning, and test out path planning methods such as RRT.
    Environment wrappers allow for flexible integration of simulated sensors to acquire point clouds or semantically segmented images.
    The use of \gls{sofa} as the underlying simulator offers a powerful and flexible platform for the development and testing of advanced control algorithms for robotic surgical systems.
    
    The goal of this software is to spur further development of \gls{rl} algorithms tailored specifically for the challenges of \gls{rals}, which include low-level control as well as a difficult exploration problem.
    The issue of exploration may be aided by the use of imitation learning with expert demonstrations, or with the help of model-based \gls{rl} methods.
    Many environments additionally require multi-agent coordination, which is a relatively little-explored topic in the field of \gls{rals}.
    Finally, \sofaenv offers an exciting benchmark for transfer learning and meta-learning approaches, since each environment can be parametrized to increase its complexity, and some environments can be modified in even more fundamental ways.
    We are convinced that improving on our baseline results through solving the challenges relating to \gls{rals} and open \gls{rl} problems will lead to a considerable leap towards cognitive surgical robotics.

\acks{
The present contribution is supported by the Helmholtz Association under the joint research school ``HIDSS4Health – Helmholtz Information and Data Science School for Health" and the Helmholtz Association's Initiative and Networking Fund on the HAICORE@KIT partition.
We would like to thank Tim Wöldecke, Viet Pham, Pit Henrich, Tom Eckardt, and Marius Steger for their help with this project.
}

\newpage

\appendix
\section*{Appendix A. Reward Functions}
\begin{tabularx}{\textwidth}{Xr}
    \toprule
    \multicolumn{1}{c}{Reward feature $\psi_i$} & Weight $w_i$ \\
    \midrule
    \multicolumn{2}{c}{ReachEnv} \\
    Distance between end effector and target & $-1.0$ \\
    Change in distance between end effector and target & $-10.0$ \\
    Time step cost & $0.0$ \\
    Action would have violated the workspace & $0.0$ \\
    Successful task & $100.0$ \\
    \multicolumn{2}{c}{DeflectSpheresEnv} \\
    Number of instruments that violate their Cartesian workspace & $0.0$ \\
    Number of instruments that violate their state limits & $0.0$ \\
    Collision between instruments & $0.0$ \\
    Distance from the instrument tip to the active sphere & $0.0$ \\
    Change in distance from the instrument tip to the active sphere & $-5.0$ \\
    Sum of the deflections of the inactive spheres & $-0.005$ \\
    Deflection of the active sphere & $0.0$ \\
    Change in deflection of the active sphere & $1.0$ \\
    Done with active sphere & $10.0$ \\
    Successful task & $100.0$ \\
    \multicolumn{2}{c}{SearchForPointEnv (without hook)} \\
    Point of interest (poi) is in the camera frame & $0.01$ \\
    Error in desired distance between camera and poi & $-0.01$ \\
    Change in error in desired distance between camera and poi & $-0.01$ \\
    Distance between the image center and the poi in the image & $-0.001$ \\
    Change in distance between the image center and the poi in the image & $-0.001$ \\
    Successful task & $100.0$ \\
    \multicolumn{2}{c}{SearchForPointEnv (with hook)} \\
    Electrocautery hook (cauter) in collision & $-0.001$ \\
    Distance between the cauter and the poi & $-0.0005$ \\
    Change in distance between the cauter and the poi & $-5.0$ \\
    Cauter touches the poi & $0.0$ \\
    Action would have violated the state limits & $0.0$ \\
    Successful task & $100.0$ \\
    \multicolumn{2}{c}{TissueManipulationEnv} \\
    Distance to target point & $-1.0$ \\
    Policy is stuck & $-5.0$ \\
    Action would have violated the workspace & $0.0$ \\
    Unstable simulation & $0.0$ \\
    Successful task & $10.0$ \\
    \multicolumn{2}{c}{PickAndPlaceEnv} \\
    Grasper is grasping the torus & $0.0$ \\
    Grasper lost its grasp on the torus &  $-30.0$ \\
    Grasper established a new grasp on the torus (only in pick phase) & $30.0$ \\
    Distance between the grasper and the torus' center of mass (only in pick phase) & $0.0$ \\
    Change in distance between the grasper and the torus' center of mass (only in pick phase) & $0.0$\\
    Distance between the grasper and point on the torus (only in pick phase) & $0.0$ \\
    Change in distance between the grasper and point on the torus (only in pick phase) & $-10.0$ \\
    Distance between the minimum lift height and torus' center of mass (only in pick phase) & $0.0$ \\
    Change in distance between the minimum lift height and torus' center of mass (only in pick phase) & $-50.0$ \\
    Distance to the active peg (only in place phase) & $0.0$ \\
    Change in distance to the active peg (only in place phase) & $-100.0$ \\
    Collision between grasper and peg & $-0.01$ \\
    Collision between grasper and board & $-0.01$ \\
    Unstable simulation & $-0.01$ \\  
    Velocity of the torus & $0.0$ \\
    Velocity of the grasper & $0.0$ \\
    Torus not on the board & $0.0$ \\
    Action would have violated the state limits & $0.0$ \\
    Action would have violated the workspace & $0.0$ \\
    Successful task & $50.0$ \\
    \multicolumn{2}{c}{GraspLiftAndTouch} \\
    Collision between cauter and grasper & $-0.1$ \\
    Collision between cauter and gallbladder & $-0.1$ \\
    Collision between cauter and liver & $-0.1$ \\
    Collision between grasper and liver & $-0.1$ \\
    Distance between cauter and target (touch phase, else) & ($-5.0$, $-0.5$) \\
    Change in distance between cauter and target & $-1.0$ \\
    Target is visible & $0.0$ \\
    Gallbladder is grasped & $20.0$ \\
    Established new grasp on gallbladder & $10.0$ \\
    Lost grasp on gallbladder & $-10.0$ \\
    Grasp contacts between grasper and gallbladder & $0.0$ \\
    Change in grasp contacts between grasper and gallbladder & $0.0$ \\
    Grasper retracts gallbladder & $0.005$ \\
    Force exerted on gallbladder & $-0.003$ \\
    Action would have violated the grasper state limits & $0.0$ \\
    Action would have violated the cauter state limits & $0.0$ \\
    Action would have violated the grasper workspace & $0.0$ \\
    Action would have violated the cauter workspace & $0.0$ \\
    Completed the active phase & $10.0$ \\
    Pushed gallbladder into liver & $-0.1$ \\
    Change in pushing gallbladder into liver & $-0.01$ \\
    Distance between grasper and gallbladder infundibulum (only in grasp phase) & $-0.2$ \\
    Change in distance between grasper and gallbladder infundibulum (only in grasp phase) & $-10.0$ \\
    Cauter activated in target (only in touch phase) & $0.0$ \\
    Change in cauter activation in target (only in touch phase) & $1.0$ \\
    Cauter touches target (only in touch phase) & $0.0$ \\
    Successful task & $200.0$ \\
    \multicolumn{2}{c}{RopeCuttingEnv} \\
    Distance to the active rope & $0.0$ \\
    Change in distance to the active rope & $-5.0$ \\
    Cut the active rope & $5.0$ \\
    Cut an inactive rope & $-5.0$ \\
    Action would have violated the state limits & $0.0$ \\
    Action would have violated the workspace & $0.0$ \\
    Failed task & $-20.0$ \\
    Successful task & $10.0$ \\
    \multicolumn{2}{c}{PrecisionCuttingEnv} \\
    Unstable simulation & $-0.0001$ \\
    Distance to cutting path & $-1.0$ \\
    Change in distance to cutting path & $-500.0$ \\
    Cut on cutting path & $0.0$ \\
    Cut outside cutting path & $-0.1$ \\
    Cut ratio of cutting path & $0.0$ \\
    Change in cut ratio of cutting path & $10.0$ \\
    Action would have violated the state limits & $0.0$ \\
    Action would have violated the workspace & $0.0$ \\
    Remote center of motion not respected & $0.0$ \\
    Successful task & $50.0$ \\
    \multicolumn{2}{c}{TissueDissectionEnv} \\
    Unstable simulation & $-1.0$ \\
    Minimal distance to connective tissue & $-10.0$ \\
    Change in minimal distance to connective tissue & $-10.0$ \\
    Cut connective tissue & $0.5$ \\
    Cut tissue flap & $-0.1$ \\
    Collision with board & $-0.1$ \\
    Action would have violated the state limits & $0.0$ \\
    Action would have violated the workspace & $0.0$ \\
    Remote center of motion not respected & $0.0$ \\
    Successful task & $50.0$ \\
    \multicolumn{2}{c}{ThreadInHoleEnv} \\
    Distance between thread tip and hole & $-0.1$ \\
    Change in distance between thread tip and hole & $-0.1$ \\
    Distance between thread center of mass and hole & $-0.0$ \\
    Change in distance between thread center of mass and hole & $-0.0$ \\
    Unstable simulation & $0.0$ \\
    Velocity of the thread & $0.0$ \\
    Velocity of the grasper & $0.0$ \\
    Action would have violated the state limits & $0.0$ \\
    Action would have violated the workspace & $0.0$ \\
    Ratio of thread in hole & $0.1$ \\
    Change of ratio of thread in hole & $1.0$ \\
    Gripper collision & $-0.1$ \\
    Successful task & $100.0$\\
    \multicolumn{2}{c}{RopeThreadingEnv} \\
    Done with active loop & $10.0$ \\
    Rope slipped out of loop & $-20.0$ \\
    Distance between rope tip and active loop & $0.0$ \\
    Change in distance to active loop & $-200.0$ \\
    Lost grasp on rope & $-0.1$ \\
    Collision between loop and grasper & $-0.1$ \\
    Collision between board and grasper & $-0.1$ \\
    Action would have violated the state limits & $0.0$ \\
    Action would have violated the workspace & $0.0$ \\
    Distance between grasper and rope (if not grasped) & $0.0$ \\
    Change in distance between grasper and rope (if not grasped) & $0.0$ \\
    Ratio of rope passed through active loop & $0.0$ \\
    Change in ratio of rope passed through active loop & $200.0$ \\
    Established grasp with both graspers (only in bimanual case) & $100.0$ \\
    Distance between second grasper and rope (only in bimanual case) & $0.0$ \\
    Change in distance between second grasper and rope (only in bimanual case) & $-200.0$ \\
    Successful task & $100.0$ \\
    \multicolumn{2}{c}{LigatingLoopEnv} \\
    Distance between loop center of mass and marking & $-0.05$ \\
    Change in distance between loop center of mass and marking & $-100.0$ \\
    Loop center of mass is inside cylinder & $0.0$ \\
    Instrument shaft not withing cylinder & $0.0$ \\
    Instrument shaft collides with cylinder & $0.0$ \\
    Overlap between loop and marking & $0.8$ \\
    Loop closes around marking & $0.5$ \\
    Loop closes in thin air & $-0.1$ \\
    Successful task & $100.0$ \\
    \bottomrule
    & \\
    \multicolumn{2}{l}{Table A: Features $\bm{\psi}$ and weights $\bm{w}$ of the environments' reward functions.}
\end{tabularx}
\newpage
\section*{Appendix B. Environment Parameters}
\begin{tabularx}{\textwidth}{lX}
    \toprule
    Environment & Parameters \\
    \midrule
    ReachEnv &
        \gls{rcm} position;
        whether to randomize the starting position;
        camera pose;
        sphere radius;
        threshold distance for task completion;
        minimum distance between target and end-effector after reset. \\
    & \\
    DeflectSpheresEnv &
        Dimensions of board and workspace;
        number of spheres;
        minimum distance between spheres;
        stiffness of the stalks;
        amount of noise applied to instrument pose on reset;
        single instrument version;
        number of deflections required for task completion;
        minimum deflection amount per sphere;
        whether to allow deflection with instrument shaft;
        whether to sample the active sphere with or without replacement of completed spheres. \\
    & \\
    SearchForPointEnv &
        Whether to add surgeon and assistant grippers to the scene;
        desired distance between target point and camera;
        threshold distance between target point and camera;
        threshold distance between target point and image center;
        threshold distance between target point and electrocautery hook;
        whether to add the electrocautery hook (active vision);
        whether the cauter has to activate in the target to complete the episode. \\
    & \\
    TissueManipulationEnv &
        Whether to randomize the grasping point;
        whether to randomize the landmark;
        camera pose;
        minimum distance between target and landmark after reset. \\
    & \\
    PickAndPlaceEnv & 
        Amount of noise applied to instrument pose on reset;
        mechanical parameters of the torus;
        number of active pegs for placing the torus;
        whether to randomize the colors of pegs and torus;
        number of points on the torus for state observations;
        whether to start with a grasped torus;
        whether to end the episode after picking is complete;
        minimum height to lift the torus for completing the pick phase;
        whether to not mark the episode as complete when the simulation is unstable. \\
    & \\
    GraspLiftAndTouch &
        Pair of start and end phase;
        threshold distance for collision checking between instruments;
        threshold distance between target point and electrocautery hook. \\
    & \\
    RopeCuttingEnv &
        Mass of the ropes;
        stiffness of the ropes;
        minimum distance between ropes;
        number of ropes;
        number of ropes to cut;
        dimensions of the walls;
        number of points per rope for state observations. \\
    & \\
    PrecisionCuttingEnv &
        Dimensions of the cloth;
        discretization of the cloth;
        stroke width of the cutting path;
        function to project onto the cloth as cutting path;
        amount of noise applied to instrument pose on reset;
        amount of noise applied to camera pose on reset;
        whether to add a grasper to the scene;
        whether to render the closest point between scissors and cutting path;
        number of points on the cloth for state observations;
        number of points on the cutting path for state observations;
        ratio of cutting path to cut for task completion;
        whether to control the scissors in Cartesian or TPSD space. \\
    & \\
    TissueDissectionEnv &
        Rows of connective tissue to cut;
        whether to add a collision model to the board;
        whether to render the closest point between instrument and connective tissue;
        amount of noise applied to instrument pose on reset;
        amount of noise applied to camera pose on reset;
        deadzone for instrument activation;
        whether to make the force that pulls back the flap controllable;
        number of points on the connective tissue for state observations;
        number of points on the flap for state observations. \\
    & \\
    ThreadInHoleEnv &
        Mechanical parameters and dimensions of the thread;
        mechanical parameters and dimensions of the hole;
        amount of noise applied to instrument pose on reset;
        amount of noise applied to hole pose on reset;
        amount of noise applied to camera pose on reset;
        number of points on the thread for state observations;
        ratio of thread to insert to complete the episode. \\
    & \\
    RopeThreadingEnv &
        Whether to randomize the graspers' poses;
        whether to start with the rope grasped;
        whether to randomize the grasp;
        number and poses of the eyelets;
        amount of noise applied to eyelet poses on reset;
        number of points on the rope for state observations;
        whether to color the eyelets according to their state;
        single grasper version;
        fraction of rope to pass through an eyelet to mark it as finished;
        whether a bimanual grasp is required to finish an eyelet. \\
    & \\
    LigatingLoopEnv &
        Whether to add a grasper to the scene;
        stiff or soft loop;
        number of points to model the loop;
        radius of the loop;
        width of the cavity marking;
        whether to randomize the marking;
        target constriction of the loop;
        target overlap between loop and marking;
        number of points on the loop for state observations;
        number of points on the cavity for state observations;
        number of points on the marking for state observations. \\
    & \\
    All environments & 
        State velocity limits to scale the normalized agent action;
        step size of a discrete action;
        observation type (RGB, RGBD, STATE). \\
    \bottomrule
    & \\
    \multicolumn{2}{l}{Table B: Environment parameters that can be used to adjust task complexity.}
\end{tabularx}

\section*{Appendix C. Simulation Parameters}
\begin{tabularx}{\textwidth}{Xlrr}
    \toprule
    Environment \hspace{4cm} & $\Delta T_s$ & $N$ & Time Limit [steps] \\
    \midrule
    ReachEnv & $0.1$ & $1$ & $500$ \\
    DeflectSpheresEnv & $0.1$ & $1$ & $500\cdot\#\text{spheres to deflect}$ \\
    SearchForPointEnv & $0.1$ & $1$ & $500$ \\
    TissueManipulationEnv & $0.1$ & $1$ & $500$ \\
    PickAndPlaceEnv & $0.05$ & $2$ & $300\cdot\#\text{phases}$ \\
    GraspLiftAndTouch & $0.05$ & $2$ & $400 + 100\cdot\#\text{phases}$ \\
    RopeCuttingEnv & $0.1$ & $1$ & max($400$, $200\cdot\#\text{ropes to cut}$) \\
    PrecisionCuttingEnv & $0.025$ & $4$ & $500$ \\
    TissueDissectionEnv & $0.01$ & $10$ & max($400$, $200\cdot\#\text{rows to cut}$) \\
    ThreadInHoleEnv & $0.01$ & $10$ & $300$ \\
    RopeThreadingEnv & $0.01$ & $10$ & $200$ \\
    LigatingLoopEnv & $0.05$ & $2$ & $500$ \\
    \bottomrule
    & \\
    \multicolumn{4}{l}{Table C: Simulation time step $\Delta T_s$, frame skip $N$, and time limit for each environment.}\\
    \multicolumn{4}{l}{Symbol \# means \textit{number of}.}
\end{tabularx}
\section*{Appendix D. State Observations}
\begin{tabularx}{\textwidth}{lrX}
    \toprule
    Environment & Size & Content \\
    \midrule
    ReachEnv & $6$ & Positions of end-effector and target. \\
    DeflectSpheresEnv & $29$ & Positions of spheres and active sphere; pose and TPSD state of instruments; id of active instrument.\\
    SearchForPointEnv & $14$/$26$ & Position of target; poses and TPSD states of camera / and hook; activation of hook. \\
    TissueManipulationEnv & $9$ & Positions of grasper, tissue point, and target point.\\
    PickAndPlaceEnv & $37$ & Pose, TPSD state, jaw angle, and grasping state of grasper; positions of $5$ points on the torus, the center of mass, and the active peg. \\
    GraspLiftAndTouch & $59$ & Poses, TPSD states, jaw angle/activation state of instruments; current phase id; positions of target and $10$ points on the gallbladder. \\
    RopeCuttingEnv & $12 + 9 \cdot (N + 1)$ & Pose, TPSD state, and activation of instrument; positions of $3$ points on all $N$ ropes and $3$ points on the active rope. \\
    PrecisionCuttingEnv & $75$ & Pose, TPSD state, and jaw angle of the scissors; positions of $10$ points on the cutting path, of $10$ points on the rest of the cloth, and of the closest point between scissors and cutting path.\\
    TissueDissectionEnv & $75$ & Pose, TPSD state, and activation of the instrument; positions of $10$ points on the flap, of $10$ points on the connective tissue, and of the closest point between instrument and connective tissue.\\\\
    ThreadInHoleEnv & $29$ & Pose and TPSD state the instrument; positions of the thread's center of mass, the hole opening, and $4$ points on the thread. \\
    RopeThreadingEnv & $50$/$63$ & Pose, TPSD state, jaw angle, and grasping state of the grasper/graspers (bimanual); pose of the active loop; positions of the rope tip and $10$ points on the rope. \\
    LigatingLoopEnv & $59$ & Pose, TPSD state, and closed ratio of the instrument; positions of $9$ points on the cylinder, $3$ points on the marking, and $6$ points on the loop. \\
    \bottomrule
    & \\
    \multicolumn{3}{l}{Table D: Size and content of the state-observations for each environment.}
\end{tabularx}


\newpage
\bibliography{references}

\end{document}